\documentclass[a4paper,10pt,onecolumn]{article}
\usepackage[english]{babel}
 \usepackage{fancyhdr} 
\usepackage[pdftex]{graphicx} 
\usepackage{caption} 
\usepackage{subcaption} 
 \usepackage{natbib} 
\usepackage{multirow} 
\usepackage{amssymb}
\usepackage[intlimits]{amsmath} 
\usepackage{amsthm} 
\usepackage[german]{nomencl}
\usepackage{booktabs} 
\usepackage{geometry}
\usepackage{bbm}
\usepackage{array}
\usepackage{listings}
\usepackage[table]{xcolor}
\usepackage{tikz}
\usetikzlibrary{shapes,arrows,shadows, positioning}
\usepackage{bm,times}
\usepackage{smartdiagram}

\definecolor{b}{rgb}{0,0,.8}	
\definecolor{g}{rgb}{0,.6,0}	
\definecolor{n}{rgb}{0,0,0}	
\definecolor{h}{rgb}{0.4,0.2,0.2}	
\definecolor{v}{rgb}{0.2,0.6,0}

\newcommand{\C}{{\mathbb C}}

\newcommand{\E}{{\mathbb E}}

\newcommand{\R}{{\mathbb R}}

\newcommand{\CC}{{\mathcal{C}}}

\newcommand{\OO}{{\mathcal{O}}}

\newcommand{\UU}{{\mathcal{U}}}

\newcommand{\XX}{{\mathcal{X}}}
\newcommand{\YY}{{\mathcal{Y}}}

\newcommand{\bsa}{\boldsymbol a}

\newcommand{\bsc}{\boldsymbol c}

\newcommand{\bsq}{\boldsymbol q}

\newcommand{\bsv}{\boldsymbol v}
\newcommand{\bsw}{\boldsymbol w}
\newcommand{\bsx}{\boldsymbol x}
\newcommand{\bsy}{\boldsymbol y}
\newcommand{\bsz}{\boldsymbol z}
\newcommand{\bsA}{\boldsymbol A}

\newcommand{\bsF}{\boldsymbol F}

\newcommand{\bsX}{\boldsymbol X}
\newcommand{\bsY}{\boldsymbol Y}
\newcommand{\bsZ}{\boldsymbol Z}
\newcommand{\bsone}{\boldsymbol 1}

\newcommand{\bsmu}{\boldsymbol \mu}

\newcommand{\bssigma}{\boldsymbol \sigma}

\newcommand{\bsvarphi}{\boldsymbol \varphi}

\newcommand{\eps}{{\varepsilon}}

\DeclareMathOperator*{\argmin}{arg\,min}

\DeclareMathOperator{\cov}{\C ov}

\newcommand{\ov}\overline

\newcommand{\rig}\right
\newcommand{\lef}\left
\newcommand{\nf}\normalfont

\definecolor{dcyan}{rgb}{0,0.5,.5}

\definecolor{dgreen}{rgb}{0,0.7,0}
 
\definecolor{dgrey}{rgb}{0.6,0.6,.6}

 \title{
The energy distance for ensemble and scenario reduction
}  

 \author{Florian Ziel}

\begin{document}
\maketitle
\lhead{\nouppercase{\leftmark}}
\begin{abstract}
Scenario reduction techniques are widely applied for solving sophisticated dynamic and stochastic programs, especially in energy and power systems, but also used in probabilistic forecasting, clustering and estimating generative adversarial networks (GANs).
We propose a new method for ensemble and scenario reduction 
based on the energy distance which is a special case of the maximum mean discrepancy (MMD). We discuss the choice of energy distance in detail, especially in comparison to
the popular Wasserstein distance which is dominating the scenario reduction literature. The energy distance is a metric between probability measures that allows for powerful tests for equality of arbitrary multivariate distributions or independence. Thanks to the latter, it is a suitable candidate for ensemble and scenario reduction problems. The theoretical properties and considered examples indicate clearly that the reduced scenario sets tend to exhibit better statistical properties for the energy distance than a corresponding reduction with respect to the Wasserstein distance.
We show applications to a Bernoulli random walk and two real data based examples for electricity demand profiles and day-ahead electricity prices.

\end{abstract}


\section{Introduction and Motivation} \label{Introduction}

In the operations research and optimization literature, ensemble and scenario reduction plays an important role for solving dynamic and stochastic programs.
There are algorithms to find an (approximately) optimal solution by valuating simulated trajectories/paths from uncertain processes which are involved in the optimization problem. 
However, subsequent optimization steps are usually very costly in terms of computational effort - therefore ensemble or scenario reduction is applied. 
This holds especially with regard to applications for 
power and energy systems, see e.g. 
\cite{growe2003scenario,di2018stochastic,biswas2019optimal,gazafroudi2019stochastic}. 
Further, reduction techniques
reduce the computational effort and memory requirements in all applications where ensembles and scenarios are relevant.
This is for instance in reporting probabilistic forecasts \cite{gneiting2014probabilistic}, estimating generative adversarial networks (esp. Wassertein-GANs and MMD-GANs)
\cite{li2017mmd,genevay2017learning}, clustering \cite{rizzo2016energy}, and visualization \cite{wang2018visualization}.

 \begin{figure}[!h]

\begin{subfigure}[b]{0.49\textwidth}
\includegraphics[width=.99\textwidth]{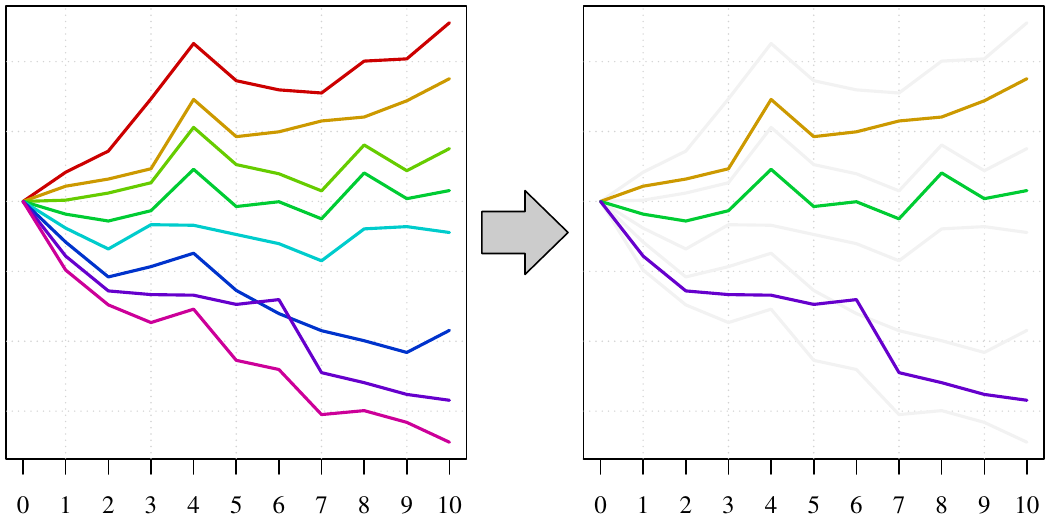} 
        \caption{Ensemble reduction}
        \label{fig_intr_ens_red}
    \end{subfigure}
    \begin{subfigure}[b]{0.49\textwidth}
\includegraphics[width=.99\textwidth]{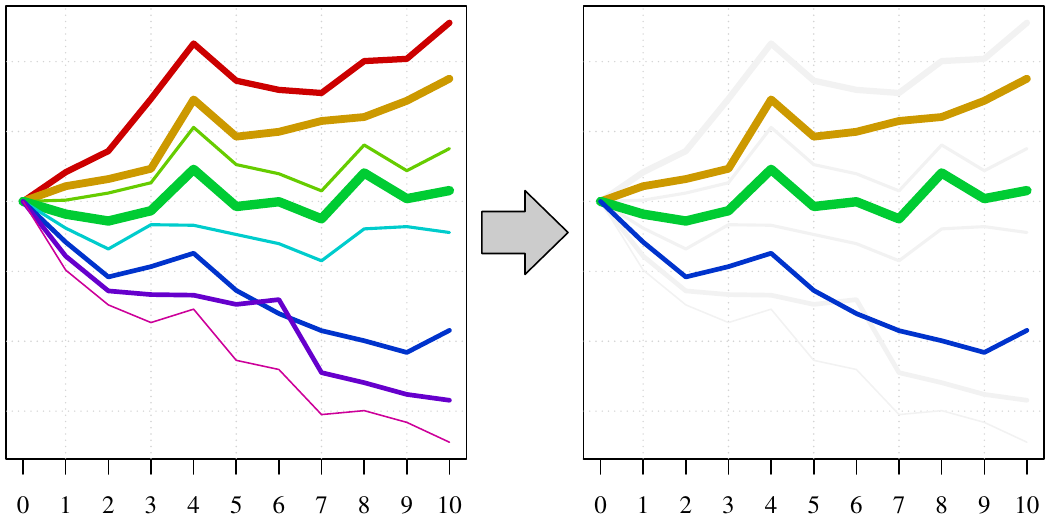} 
        \caption{Weighted ensemble reduction}
        \label{fig_intr_wens_red}
    \end{subfigure}

    \begin{subfigure}[b]{0.49\textwidth}
\includegraphics[width=.99\textwidth]{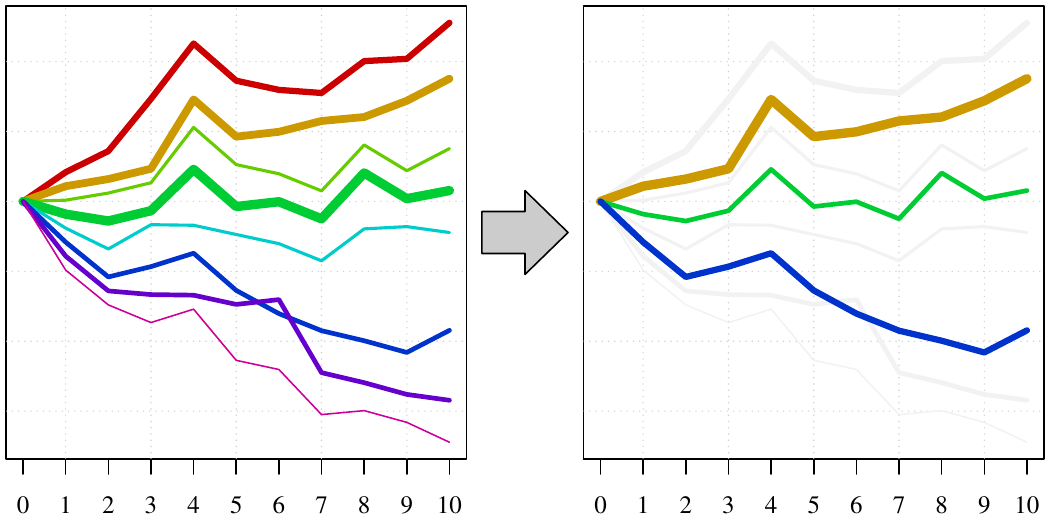} 
        \caption{(Discrete) scenario reduction}
        \label{fig_intr_scen_red}
    \end{subfigure}
    \begin{subfigure}[b]{0.49\textwidth}
\includegraphics[width=.99\textwidth]{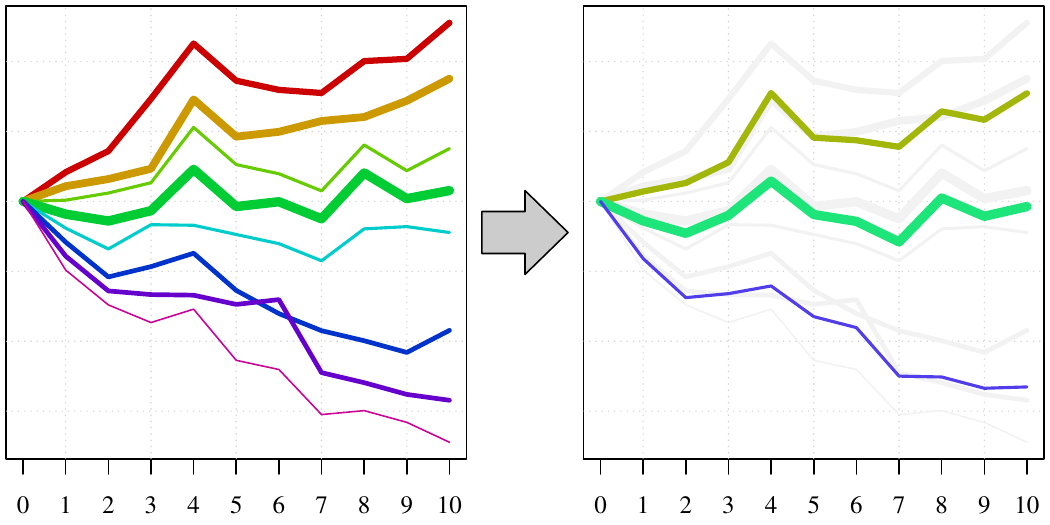} 
        \caption{Continuous scenario reduction}
        \label{fig_intr_cscen_red}
    \end{subfigure}
    
 \caption{Illustration of ensemble and scenario reduction principles. }
 \label{fig_intro}
 \end{figure}

Based on a (weighted) set of simulated trajectories or paths, a reduction technique is applied.
The target for \emph{ensemble reduction} is to find an \emph{optimal} subset of $m$ paths out of the simulated ensemble set with $n$ paths, where $m<n$, see Fig. \ref{fig_intr_ens_red} and \ref{fig_intr_wens_red} 
for illustration. Thus, the weight distribution of the scenarios remain unchanged. In the equally weighted case (Fig. \ref{fig_intr_ens_red}) all
trajectories in the reduced ensemble receive a weight of $1/m$, in Fig. \ref{fig_intr_wens_red}, the remaining weights are scaled to sum up to 1.
Alongside with the classification also used in \cite{rujeerapaiboon2018scenario}, in (discrete) \emph{scenario reduction} (compare Fig. \ref{fig_intr_scen_red}) we aim to find not only the $m$ subset paths but also the \emph{optimal} associated probabilities. 
Thus, scenario reduction is usually a more sophisticated problem than ensemble reduction. Even more advanced and related to clustering methods is \emph{continuous scenario reduction} (also known as scenario generation, see Fig. \ref{fig_intr_cscen_red}). There we are looking for $m$ new paths
and weights that approximate the target distribution well. Which method is required depends a lot on the underlying problem. In 
this manuscript, we do not focus on continuous scenario reduction. Whenever we talk about scenario reduction, we mean discrete scenario reduction.

Obviously, the crucial question is: What means \emph{optimal} in our context? In the ensemble and scenario reduction literature, different 
criteria for optimality are suggested. An important selection criterion is the choice of a distance (or metric) which characterizes to  a
certain extent how close the $m$ selected trajectories are to the considered large set of $n$ trajectories in the scenario set. 
The vast majority of articles favor the Wasserstein type distances resp. metrics (also Fortet-Mourier distance Kantorovich/Kantorovich-Rubinstein distance, earth mover distance, optimal transport distance) for scenario reduction,  see \cite{rujeerapaiboon2018scenario,glanzer2020multiscale}. This also holds for all applications mentioned in the first paragraph. 
There are two main reasons: On the one hand,
there are useful stability results in stochastic programming for the Wasserstein metric available. They guarantee under mild assumptions on the stochastic programme that a scenario reduction that approximates the full scenario set well leads to close to optimal solutions of the stochastic optimization problem, see e.g. \cite{romisch2003stability, rujeerapaiboon2018scenario} for more details.
On the other hand there is the fast 
scenario reduction method proposed by \cite{dupavcova2003scenario}, along with the implementation in the
General Algebraic Modeling System (GAMS) by the software \texttt{Scenred}. It is based on the Wasserstein distance and the optimal redistribution rule (see e.g. \cite{growe2003scenario}). It represents the reduction problem as a (mass) transportation problem which has an explicit solution for the scenario reduction problem.  
As the Wasserstein distance is a metric for probability measure that characterizes weak convergence of measures, it seems to be a suitable candidate for reduction problems. For instance \cite{rujeerapaiboon2018scenario} states: \emph{'The modern stability theory of stochastic programming indicates that the distance may serve as a natural candidate for this probability metric.'} in favor for the Wasserstein distance for scenario reduction problems.
However, there are reduction techniques based on other distances, such as the discrepancy distances \cite{henrion2009scenario} and methods
which are not based on probability distances, e.g. reduction techniques based on Euclidean distances, sampling or clustering
\cite{keko2015impact, park2019comparing, zhou2019improved}. 

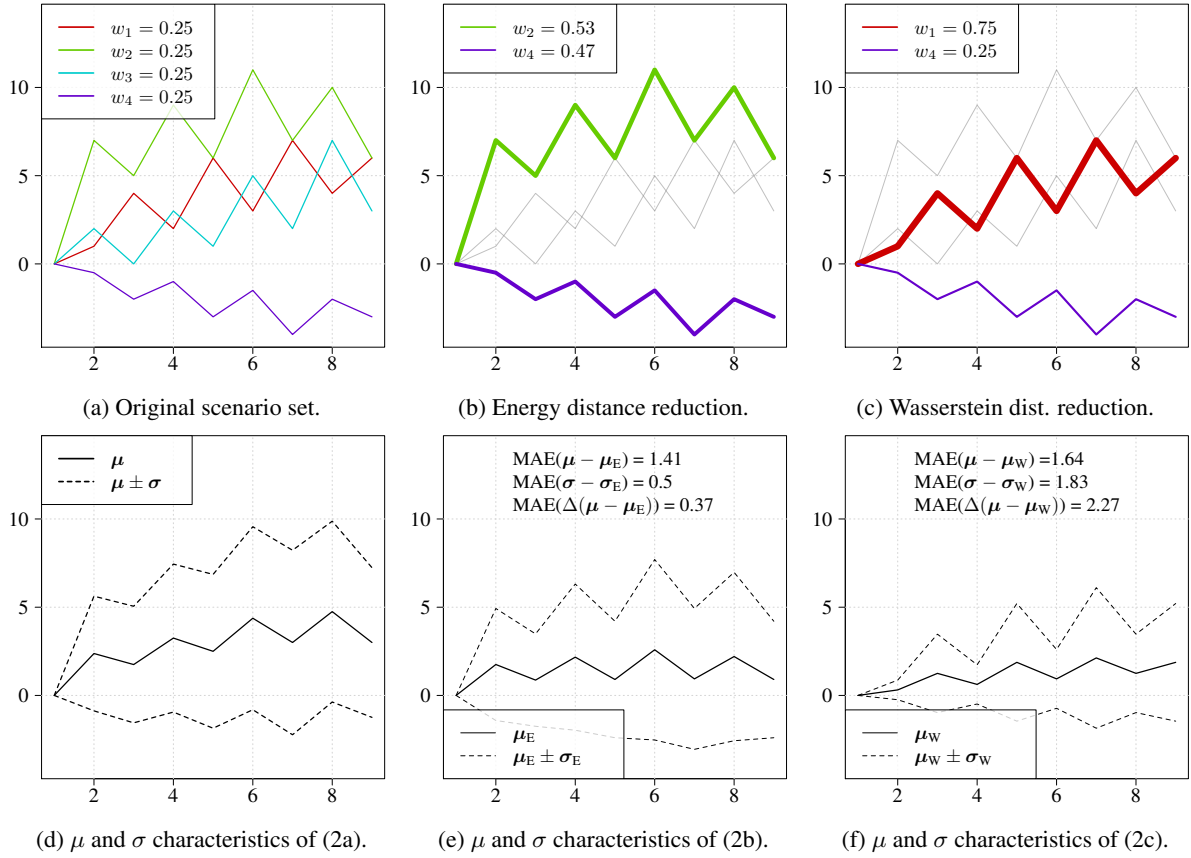
\begin{figure}

\begin{subfigure}[b]{0.325\textwidth}
\centering
\resizebox{.999\textwidth}{!}{ 
\begin{tikzpicture}[x=1pt,y=1pt]
\definecolor{fillColor}{RGB}{255,255,255}
\path[use as bounding box,fill=fillColor,fill opacity=0.00] (0,0) rectangle (361.35,361.35);
\begin{scope}
\path[clip] ( 32.40, 30.00) rectangle (352.95,352.95);
\definecolor{drawColor}{RGB}{255,255,255}

\path[draw=drawColor,line width= 0.4pt,line join=round,line cap=round] ( 44.27,125.02) circle (  2.25);

\path[draw=drawColor,line width= 0.4pt,line join=round,line cap=round] ( 81.37,141.64) circle (  2.25);

\path[draw=drawColor,line width= 0.4pt,line join=round,line cap=round] (118.47,158.25) circle (  2.25);

\path[draw=drawColor,line width= 0.4pt,line join=round,line cap=round] (155.57,174.86) circle (  2.25);

\path[draw=drawColor,line width= 0.4pt,line join=round,line cap=round] (192.67,191.47) circle (  2.25);

\path[draw=drawColor,line width= 0.4pt,line join=round,line cap=round] (229.78,208.09) circle (  2.25);

\path[draw=drawColor,line width= 0.4pt,line join=round,line cap=round] (266.88,224.70) circle (  2.25);

\path[draw=drawColor,line width= 0.4pt,line join=round,line cap=round] (303.98,241.31) circle (  2.25);

\path[draw=drawColor,line width= 0.4pt,line join=round,line cap=round] (341.08,257.93) circle (  2.25);
\end{scope}
\begin{scope}
\path[clip] (  0.00,  0.00) rectangle (361.35,361.35);
\definecolor{drawColor}{RGB}{0,0,0}

\path[draw=drawColor,line width= 0.4pt,line join=round,line cap=round] ( 81.37, 30.00) -- (303.98, 30.00);

\path[draw=drawColor,line width= 0.4pt,line join=round,line cap=round] ( 81.37, 30.00) -- ( 81.37, 24.00);

\path[draw=drawColor,line width= 0.4pt,line join=round,line cap=round] (155.57, 30.00) -- (155.57, 24.00);

\path[draw=drawColor,line width= 0.4pt,line join=round,line cap=round] (229.78, 30.00) -- (229.78, 24.00);

\path[draw=drawColor,line width= 0.4pt,line join=round,line cap=round] (303.98, 30.00) -- (303.98, 24.00);

\node[text=drawColor,anchor=base,inner sep=0pt, outer sep=0pt, scale=  1.90] at ( 81.37,  8.40) {2};

\node[text=drawColor,anchor=base,inner sep=0pt, outer sep=0pt, scale=  1.90] at (155.57,  8.40) {4};

\node[text=drawColor,anchor=base,inner sep=0pt, outer sep=0pt, scale=  1.90] at (229.78,  8.40) {6};

\node[text=drawColor,anchor=base,inner sep=0pt, outer sep=0pt, scale=  1.90] at (303.98,  8.40) {8};

\path[draw=drawColor,line width= 0.4pt,line join=round,line cap=round] ( 32.40,108.41) -- ( 32.40,274.54);

\path[draw=drawColor,line width= 0.4pt,line join=round,line cap=round] ( 32.40,108.41) -- ( 26.40,108.41);

\path[draw=drawColor,line width= 0.4pt,line join=round,line cap=round] ( 32.40,191.47) -- ( 26.40,191.47);

\path[draw=drawColor,line width= 0.4pt,line join=round,line cap=round] ( 32.40,274.54) -- ( 26.40,274.54);

\node[text=drawColor,anchor=base east,inner sep=0pt, outer sep=0pt, scale=  1.90] at ( 20.40,101.87) {0};

\node[text=drawColor,anchor=base east,inner sep=0pt, outer sep=0pt, scale=  1.90] at ( 20.40,184.93) {5};

\node[text=drawColor,anchor=base east,inner sep=0pt, outer sep=0pt, scale=  1.90] at ( 20.40,268.00) {10};

\path[draw=drawColor,line width= 0.4pt,line join=round,line cap=round] ( 32.40, 30.00) --
	(352.95, 30.00) --
	(352.95,352.95) --
	( 32.40,352.95) --
	( 32.40, 30.00);
\end{scope}
\begin{scope}
\path[clip] ( 32.40, 30.00) rectangle (352.95,352.95);
\definecolor{drawColor}{RGB}{211,211,211}

\path[draw=drawColor,line width= 0.4pt,dash pattern=on 1pt off 3pt ,line join=round,line cap=round] ( 81.37, 30.00) -- ( 81.37,352.95);

\path[draw=drawColor,line width= 0.4pt,dash pattern=on 1pt off 3pt ,line join=round,line cap=round] (155.57, 30.00) -- (155.57,352.95);

\path[draw=drawColor,line width= 0.4pt,dash pattern=on 1pt off 3pt ,line join=round,line cap=round] (229.78, 30.00) -- (229.78,352.95);

\path[draw=drawColor,line width= 0.4pt,dash pattern=on 1pt off 3pt ,line join=round,line cap=round] (303.98, 30.00) -- (303.98,352.95);

\path[draw=drawColor,line width= 0.4pt,dash pattern=on 1pt off 3pt ,line join=round,line cap=round] ( 32.40,108.41) -- (352.95,108.41);

\path[draw=drawColor,line width= 0.4pt,dash pattern=on 1pt off 3pt ,line join=round,line cap=round] ( 32.40,191.47) -- (352.95,191.47);

\path[draw=drawColor,line width= 0.4pt,dash pattern=on 1pt off 3pt ,line join=round,line cap=round] ( 32.40,274.54) -- (352.95,274.54);
\definecolor{drawColor}{RGB}{204,0,0}

\path[draw=drawColor,line width= 1.2pt,line join=round,line cap=round] ( 44.27,108.41) --
	( 81.37,125.02) --
	(118.47,174.86) --
	(155.57,141.64) --
	(192.67,208.09) --
	(229.78,158.25) --
	(266.88,224.70) --
	(303.98,174.86) --
	(341.08,208.09);
\definecolor{drawColor}{RGB}{102,204,0}

\path[draw=drawColor,line width= 1.2pt,line join=round,line cap=round] ( 44.27,108.41) --
	( 81.37,224.70) --
	(118.47,191.47) --
	(155.57,257.93) --
	(192.67,208.09) --
	(229.78,291.15) --
	(266.88,224.70) --
	(303.98,274.54) --
	(341.08,208.09);
\definecolor{drawColor}{RGB}{0,204,204}

\path[draw=drawColor,line width= 1.2pt,line join=round,line cap=round] ( 44.27,108.41) --
	( 81.37,141.64) --
	(118.47,108.41) --
	(155.57,158.25) --
	(192.67,125.02) --
	(229.78,191.47) --
	(266.88,141.64) --
	(303.98,224.70) --
	(341.08,158.25);
\definecolor{drawColor}{RGB}{102,0,204}

\path[draw=drawColor,line width= 1.2pt,line join=round,line cap=round] ( 44.27,108.41) --
	( 81.37,100.11) --
	(118.47, 75.19) --
	(155.57, 91.80) --
	(192.67, 58.57) --
	(229.78, 83.49) --
	(266.88, 41.96) --
	(303.98, 75.19) --
	(341.08, 58.57);
\definecolor{drawColor}{RGB}{0,0,0}
\definecolor{fillColor}{RGB}{255,255,255}

\path[draw=drawColor,line width= 0.4pt,line join=round,line cap=round,fill=fillColor,fill opacity=0.80] ( 32.40,352.95) rectangle (194.26,244.95);
\definecolor{drawColor}{RGB}{204,0,0}

\path[draw=drawColor,line width= 1.6pt,line join=round,line cap=round] ( 48.60,331.35) -- ( 81.00,331.35);
\definecolor{drawColor}{RGB}{102,204,0}

\path[draw=drawColor,line width= 1.6pt,line join=round,line cap=round] ( 48.60,309.75) -- ( 81.00,309.75);
\definecolor{drawColor}{RGB}{0,204,204}

\path[draw=drawColor,line width= 1.6pt,line join=round,line cap=round] ( 48.60,288.15) -- ( 81.00,288.15);
\definecolor{drawColor}{RGB}{102,0,204}

\path[draw=drawColor,line width= 1.6pt,line join=round,line cap=round] ( 48.60,266.55) -- ( 81.00,266.55);
\definecolor{drawColor}{RGB}{0,0,0}

\node[text=drawColor,anchor=base west,inner sep=0pt, outer sep=0pt, scale=  1.80] at ( 97.20,325.15) {$w_{1} = 0.25$  \textcolor{white}{-}};

\node[text=drawColor,anchor=base west,inner sep=0pt, outer sep=0pt, scale=  1.80] at ( 97.20,303.55) {$w_{2} = 0.25$  \textcolor{white}{-}};

\node[text=drawColor,anchor=base west,inner sep=0pt, outer sep=0pt, scale=  1.80] at ( 97.20,281.95) {$w_{3} = 0.25$  \textcolor{white}{-}};

\node[text=drawColor,anchor=base west,inner sep=0pt, outer sep=0pt, scale=  1.80] at ( 97.20,260.35) {$w_{4} = 0.25$  \textcolor{white}{-}};
\end{scope}
\end{tikzpicture}  }
        \caption{Original scenario set.}
        \label{fig_toy_a}
    \end{subfigure}
\begin{subfigure}[b]{0.325\textwidth}
\centering
\resizebox{.999\textwidth}{!}{ 
\begin{tikzpicture}[x=1pt,y=1pt]
\definecolor{fillColor}{RGB}{255,255,255}
\path[use as bounding box,fill=fillColor,fill opacity=0.00] (0,0) rectangle (361.35,361.35);
\begin{scope}
\path[clip] ( 32.40, 30.00) rectangle (352.95,352.95);
\definecolor{drawColor}{RGB}{255,255,255}

\path[draw=drawColor,line width= 0.4pt,line join=round,line cap=round] ( 44.27,125.02) circle (  2.25);

\path[draw=drawColor,line width= 0.4pt,line join=round,line cap=round] ( 81.37,141.64) circle (  2.25);

\path[draw=drawColor,line width= 0.4pt,line join=round,line cap=round] (118.47,158.25) circle (  2.25);

\path[draw=drawColor,line width= 0.4pt,line join=round,line cap=round] (155.57,174.86) circle (  2.25);

\path[draw=drawColor,line width= 0.4pt,line join=round,line cap=round] (192.67,191.47) circle (  2.25);

\path[draw=drawColor,line width= 0.4pt,line join=round,line cap=round] (229.78,208.09) circle (  2.25);

\path[draw=drawColor,line width= 0.4pt,line join=round,line cap=round] (266.88,224.70) circle (  2.25);

\path[draw=drawColor,line width= 0.4pt,line join=round,line cap=round] (303.98,241.31) circle (  2.25);

\path[draw=drawColor,line width= 0.4pt,line join=round,line cap=round] (341.08,257.93) circle (  2.25);
\end{scope}
\begin{scope}
\path[clip] (  0.00,  0.00) rectangle (361.35,361.35);
\definecolor{drawColor}{RGB}{0,0,0}

\path[draw=drawColor,line width= 0.4pt,line join=round,line cap=round] ( 81.37, 30.00) -- (303.98, 30.00);

\path[draw=drawColor,line width= 0.4pt,line join=round,line cap=round] ( 81.37, 30.00) -- ( 81.37, 24.00);

\path[draw=drawColor,line width= 0.4pt,line join=round,line cap=round] (155.57, 30.00) -- (155.57, 24.00);

\path[draw=drawColor,line width= 0.4pt,line join=round,line cap=round] (229.78, 30.00) -- (229.78, 24.00);

\path[draw=drawColor,line width= 0.4pt,line join=round,line cap=round] (303.98, 30.00) -- (303.98, 24.00);

\node[text=drawColor,anchor=base,inner sep=0pt, outer sep=0pt, scale=  1.90] at ( 81.37,  8.40) {2};

\node[text=drawColor,anchor=base,inner sep=0pt, outer sep=0pt, scale=  1.90] at (155.57,  8.40) {4};

\node[text=drawColor,anchor=base,inner sep=0pt, outer sep=0pt, scale=  1.90] at (229.78,  8.40) {6};

\node[text=drawColor,anchor=base,inner sep=0pt, outer sep=0pt, scale=  1.90] at (303.98,  8.40) {8};

\path[draw=drawColor,line width= 0.4pt,line join=round,line cap=round] ( 32.40,108.41) -- ( 32.40,274.54);

\path[draw=drawColor,line width= 0.4pt,line join=round,line cap=round] ( 32.40,108.41) -- ( 26.40,108.41);

\path[draw=drawColor,line width= 0.4pt,line join=round,line cap=round] ( 32.40,191.47) -- ( 26.40,191.47);

\path[draw=drawColor,line width= 0.4pt,line join=round,line cap=round] ( 32.40,274.54) -- ( 26.40,274.54);

\node[text=drawColor,anchor=base east,inner sep=0pt, outer sep=0pt, scale=  1.90] at ( 20.40,101.87) {0};

\node[text=drawColor,anchor=base east,inner sep=0pt, outer sep=0pt, scale=  1.90] at ( 20.40,184.93) {5};

\node[text=drawColor,anchor=base east,inner sep=0pt, outer sep=0pt, scale=  1.90] at ( 20.40,268.00) {10};

\path[draw=drawColor,line width= 0.4pt,line join=round,line cap=round] ( 32.40, 30.00) --
	(352.95, 30.00) --
	(352.95,352.95) --
	( 32.40,352.95) --
	( 32.40, 30.00);
\end{scope}
\begin{scope}
\path[clip] ( 32.40, 30.00) rectangle (352.95,352.95);
\definecolor{drawColor}{RGB}{211,211,211}

\path[draw=drawColor,line width= 0.4pt,dash pattern=on 1pt off 3pt ,line join=round,line cap=round] ( 81.37, 30.00) -- ( 81.37,352.95);

\path[draw=drawColor,line width= 0.4pt,dash pattern=on 1pt off 3pt ,line join=round,line cap=round] (155.57, 30.00) -- (155.57,352.95);

\path[draw=drawColor,line width= 0.4pt,dash pattern=on 1pt off 3pt ,line join=round,line cap=round] (229.78, 30.00) -- (229.78,352.95);

\path[draw=drawColor,line width= 0.4pt,dash pattern=on 1pt off 3pt ,line join=round,line cap=round] (303.98, 30.00) -- (303.98,352.95);

\path[draw=drawColor,line width= 0.4pt,dash pattern=on 1pt off 3pt ,line join=round,line cap=round] ( 32.40,108.41) -- (352.95,108.41);

\path[draw=drawColor,line width= 0.4pt,dash pattern=on 1pt off 3pt ,line join=round,line cap=round] ( 32.40,191.47) -- (352.95,191.47);

\path[draw=drawColor,line width= 0.4pt,dash pattern=on 1pt off 3pt ,line join=round,line cap=round] ( 32.40,274.54) -- (352.95,274.54);
\definecolor{drawColor}{RGB}{128,128,128}

\path[draw=drawColor,draw opacity=0.50,line width= 0.8pt,line join=round,line cap=round] ( 44.27,108.41) --
	( 81.37,125.02) --
	(118.47,174.86) --
	(155.57,141.64) --
	(192.67,208.09) --
	(229.78,158.25) --
	(266.88,224.70) --
	(303.98,174.86) --
	(341.08,208.09);

\path[draw=drawColor,draw opacity=0.50,line width= 0.8pt,line join=round,line cap=round] ( 44.27,108.41) --
	( 81.37,224.70) --
	(118.47,191.47) --
	(155.57,257.93) --
	(192.67,208.09) --
	(229.78,291.15) --
	(266.88,224.70) --
	(303.98,274.54) --
	(341.08,208.09);

\path[draw=drawColor,draw opacity=0.50,line width= 0.8pt,line join=round,line cap=round] ( 44.27,108.41) --
	( 81.37,141.64) --
	(118.47,108.41) --
	(155.57,158.25) --
	(192.67,125.02) --
	(229.78,191.47) --
	(266.88,141.64) --
	(303.98,224.70) --
	(341.08,158.25);

\path[draw=drawColor,draw opacity=0.50,line width= 0.8pt,line join=round,line cap=round] ( 44.27,108.41) --
	( 81.37,100.11) --
	(118.47, 75.19) --
	(155.57, 91.80) --
	(192.67, 58.57) --
	(229.78, 83.49) --
	(266.88, 41.96) --
	(303.98, 75.19) --
	(341.08, 58.57);
\definecolor{drawColor}{RGB}{102,204,0}

\path[draw=drawColor,line width= 4.3pt,line join=round,line cap=round] ( 44.27,108.41) --
	( 81.37,224.70) --
	(118.47,191.47) --
	(155.57,257.93) --
	(192.67,208.09) --
	(229.78,291.15) --
	(266.88,224.70) --
	(303.98,274.54) --
	(341.08,208.09);
\definecolor{drawColor}{RGB}{102,0,204}

\path[draw=drawColor,line width= 3.7pt,line join=round,line cap=round] ( 44.27,108.41) --
	( 81.37,100.11) --
	(118.47, 75.19) --
	(155.57, 91.80) --
	(192.67, 58.57) --
	(229.78, 83.49) --
	(266.88, 41.96) --
	(303.98, 75.19) --
	(341.08, 58.57);
\definecolor{drawColor}{RGB}{0,0,0}
\definecolor{fillColor}{RGB}{255,255,255}

\path[draw=drawColor,line width= 0.4pt,line join=round,line cap=round,fill=fillColor,fill opacity=0.80] ( 32.40,352.95) rectangle (194.26,288.15);
\definecolor{drawColor}{RGB}{102,204,0}

\path[draw=drawColor,line width= 1.6pt,line join=round,line cap=round] ( 48.60,331.35) -- ( 81.00,331.35);
\definecolor{drawColor}{RGB}{102,0,204}

\path[draw=drawColor,line width= 1.6pt,line join=round,line cap=round] ( 48.60,309.75) -- ( 81.00,309.75);
\definecolor{drawColor}{RGB}{0,0,0}

\node[text=drawColor,anchor=base west,inner sep=0pt, outer sep=0pt, scale=  1.80] at ( 97.20,325.15) {$w_{2} = 0.53$  \textcolor{white}{-}};

\node[text=drawColor,anchor=base west,inner sep=0pt, outer sep=0pt, scale=  1.80] at ( 97.20,303.55) {$w_{4} = 0.47$  \textcolor{white}{-}};
\end{scope}
\end{tikzpicture}  }
        \caption{Energy distance reduction.}
        \label{fig_toy_b}
    \end{subfigure}
\begin{subfigure}[b]{0.325\textwidth}
\centering
\resizebox{.999\textwidth}{!}{ 
\begin{tikzpicture}[x=1pt,y=1pt]
\definecolor{fillColor}{RGB}{255,255,255}
\path[use as bounding box,fill=fillColor,fill opacity=0.00] (0,0) rectangle (361.35,361.35);
\begin{scope}
\path[clip] ( 32.40, 30.00) rectangle (352.95,352.95);
\definecolor{drawColor}{RGB}{255,255,255}

\path[draw=drawColor,line width= 0.4pt,line join=round,line cap=round] ( 44.27,125.02) circle (  2.25);

\path[draw=drawColor,line width= 0.4pt,line join=round,line cap=round] ( 81.37,141.64) circle (  2.25);

\path[draw=drawColor,line width= 0.4pt,line join=round,line cap=round] (118.47,158.25) circle (  2.25);

\path[draw=drawColor,line width= 0.4pt,line join=round,line cap=round] (155.57,174.86) circle (  2.25);

\path[draw=drawColor,line width= 0.4pt,line join=round,line cap=round] (192.67,191.47) circle (  2.25);

\path[draw=drawColor,line width= 0.4pt,line join=round,line cap=round] (229.78,208.09) circle (  2.25);

\path[draw=drawColor,line width= 0.4pt,line join=round,line cap=round] (266.88,224.70) circle (  2.25);

\path[draw=drawColor,line width= 0.4pt,line join=round,line cap=round] (303.98,241.31) circle (  2.25);

\path[draw=drawColor,line width= 0.4pt,line join=round,line cap=round] (341.08,257.93) circle (  2.25);
\end{scope}
\begin{scope}
\path[clip] (  0.00,  0.00) rectangle (361.35,361.35);
\definecolor{drawColor}{RGB}{0,0,0}

\path[draw=drawColor,line width= 0.4pt,line join=round,line cap=round] ( 81.37, 30.00) -- (303.98, 30.00);

\path[draw=drawColor,line width= 0.4pt,line join=round,line cap=round] ( 81.37, 30.00) -- ( 81.37, 24.00);

\path[draw=drawColor,line width= 0.4pt,line join=round,line cap=round] (155.57, 30.00) -- (155.57, 24.00);

\path[draw=drawColor,line width= 0.4pt,line join=round,line cap=round] (229.78, 30.00) -- (229.78, 24.00);

\path[draw=drawColor,line width= 0.4pt,line join=round,line cap=round] (303.98, 30.00) -- (303.98, 24.00);

\node[text=drawColor,anchor=base,inner sep=0pt, outer sep=0pt, scale=  1.90] at ( 81.37,  8.40) {2};

\node[text=drawColor,anchor=base,inner sep=0pt, outer sep=0pt, scale=  1.90] at (155.57,  8.40) {4};

\node[text=drawColor,anchor=base,inner sep=0pt, outer sep=0pt, scale=  1.90] at (229.78,  8.40) {6};

\node[text=drawColor,anchor=base,inner sep=0pt, outer sep=0pt, scale=  1.90] at (303.98,  8.40) {8};

\path[draw=drawColor,line width= 0.4pt,line join=round,line cap=round] ( 32.40,108.41) -- ( 32.40,274.54);

\path[draw=drawColor,line width= 0.4pt,line join=round,line cap=round] ( 32.40,108.41) -- ( 26.40,108.41);

\path[draw=drawColor,line width= 0.4pt,line join=round,line cap=round] ( 32.40,191.47) -- ( 26.40,191.47);

\path[draw=drawColor,line width= 0.4pt,line join=round,line cap=round] ( 32.40,274.54) -- ( 26.40,274.54);

\node[text=drawColor,anchor=base east,inner sep=0pt, outer sep=0pt, scale=  1.90] at ( 20.40,101.87) {0};

\node[text=drawColor,anchor=base east,inner sep=0pt, outer sep=0pt, scale=  1.90] at ( 20.40,184.93) {5};

\node[text=drawColor,anchor=base east,inner sep=0pt, outer sep=0pt, scale=  1.90] at ( 20.40,268.00) {10};

\path[draw=drawColor,line width= 0.4pt,line join=round,line cap=round] ( 32.40, 30.00) --
	(352.95, 30.00) --
	(352.95,352.95) --
	( 32.40,352.95) --
	( 32.40, 30.00);
\end{scope}
\begin{scope}
\path[clip] ( 32.40, 30.00) rectangle (352.95,352.95);
\definecolor{drawColor}{RGB}{211,211,211}

\path[draw=drawColor,line width= 0.4pt,dash pattern=on 1pt off 3pt ,line join=round,line cap=round] ( 81.37, 30.00) -- ( 81.37,352.95);

\path[draw=drawColor,line width= 0.4pt,dash pattern=on 1pt off 3pt ,line join=round,line cap=round] (155.57, 30.00) -- (155.57,352.95);

\path[draw=drawColor,line width= 0.4pt,dash pattern=on 1pt off 3pt ,line join=round,line cap=round] (229.78, 30.00) -- (229.78,352.95);

\path[draw=drawColor,line width= 0.4pt,dash pattern=on 1pt off 3pt ,line join=round,line cap=round] (303.98, 30.00) -- (303.98,352.95);

\path[draw=drawColor,line width= 0.4pt,dash pattern=on 1pt off 3pt ,line join=round,line cap=round] ( 32.40,108.41) -- (352.95,108.41);

\path[draw=drawColor,line width= 0.4pt,dash pattern=on 1pt off 3pt ,line join=round,line cap=round] ( 32.40,191.47) -- (352.95,191.47);

\path[draw=drawColor,line width= 0.4pt,dash pattern=on 1pt off 3pt ,line join=round,line cap=round] ( 32.40,274.54) -- (352.95,274.54);
\definecolor{drawColor}{RGB}{128,128,128}

\path[draw=drawColor,draw opacity=0.50,line width= 0.8pt,line join=round,line cap=round] ( 44.27,108.41) --
	( 81.37,125.02) --
	(118.47,174.86) --
	(155.57,141.64) --
	(192.67,208.09) --
	(229.78,158.25) --
	(266.88,224.70) --
	(303.98,174.86) --
	(341.08,208.09);

\path[draw=drawColor,draw opacity=0.50,line width= 0.8pt,line join=round,line cap=round] ( 44.27,108.41) --
	( 81.37,224.70) --
	(118.47,191.47) --
	(155.57,257.93) --
	(192.67,208.09) --
	(229.78,291.15) --
	(266.88,224.70) --
	(303.98,274.54) --
	(341.08,208.09);

\path[draw=drawColor,draw opacity=0.50,line width= 0.8pt,line join=round,line cap=round] ( 44.27,108.41) --
	( 81.37,141.64) --
	(118.47,108.41) --
	(155.57,158.25) --
	(192.67,125.02) --
	(229.78,191.47) --
	(266.88,141.64) --
	(303.98,224.70) --
	(341.08,158.25);

\path[draw=drawColor,draw opacity=0.50,line width= 0.8pt,line join=round,line cap=round] ( 44.27,108.41) --
	( 81.37,100.11) --
	(118.47, 75.19) --
	(155.57, 91.80) --
	(192.67, 58.57) --
	(229.78, 83.49) --
	(266.88, 41.96) --
	(303.98, 75.19) --
	(341.08, 58.57);
\definecolor{drawColor}{RGB}{204,0,0}

\path[draw=drawColor,line width= 6.0pt,line join=round,line cap=round] ( 44.27,108.41) --
	( 81.37,125.02) --
	(118.47,174.86) --
	(155.57,141.64) --
	(192.67,208.09) --
	(229.78,158.25) --
	(266.88,224.70) --
	(303.98,174.86) --
	(341.08,208.09);
\definecolor{drawColor}{RGB}{102,0,204}

\path[draw=drawColor,line width= 2.0pt,line join=round,line cap=round] ( 44.27,108.41) --
	( 81.37,100.11) --
	(118.47, 75.19) --
	(155.57, 91.80) --
	(192.67, 58.57) --
	(229.78, 83.49) --
	(266.88, 41.96) --
	(303.98, 75.19) --
	(341.08, 58.57);
\definecolor{drawColor}{RGB}{0,0,0}
\definecolor{fillColor}{RGB}{255,255,255}

\path[draw=drawColor,line width= 0.4pt,line join=round,line cap=round,fill=fillColor,fill opacity=0.80] ( 32.40,352.95) rectangle (194.26,288.15);
\definecolor{drawColor}{RGB}{204,0,0}

\path[draw=drawColor,line width= 1.6pt,line join=round,line cap=round] ( 48.60,331.35) -- ( 81.00,331.35);
\definecolor{drawColor}{RGB}{102,0,204}

\path[draw=drawColor,line width= 1.6pt,line join=round,line cap=round] ( 48.60,309.75) -- ( 81.00,309.75);
\definecolor{drawColor}{RGB}{0,0,0}

\node[text=drawColor,anchor=base west,inner sep=0pt, outer sep=0pt, scale=  1.80] at ( 97.20,325.15) {$w_{1} = 0.75$  \textcolor{white}{-}};

\node[text=drawColor,anchor=base west,inner sep=0pt, outer sep=0pt, scale=  1.80] at ( 97.20,303.55) {$w_{4} = 0.25$  \textcolor{white}{-}};
\end{scope}
\end{tikzpicture}  }
        \caption{Wasserstein dist. reduction.}
        \label{fig_toy_c}
    \end{subfigure}
    
\begin{subfigure}[b]{0.325\textwidth}
\centering
\resizebox{.999\textwidth}{!}{ 
\begin{tikzpicture}[x=1pt,y=1pt]
\definecolor{fillColor}{RGB}{255,255,255}
\path[use as bounding box,fill=fillColor,fill opacity=0.00] (0,0) rectangle (361.35,361.35);
\begin{scope}
\path[clip] ( 32.40, 30.00) rectangle (352.95,352.95);
\definecolor{drawColor}{RGB}{255,255,255}

\path[draw=drawColor,line width= 0.4pt,line join=round,line cap=round] ( 44.27,125.02) circle (  2.25);

\path[draw=drawColor,line width= 0.4pt,line join=round,line cap=round] ( 81.37,141.64) circle (  2.25);

\path[draw=drawColor,line width= 0.4pt,line join=round,line cap=round] (118.47,158.25) circle (  2.25);

\path[draw=drawColor,line width= 0.4pt,line join=round,line cap=round] (155.57,174.86) circle (  2.25);

\path[draw=drawColor,line width= 0.4pt,line join=round,line cap=round] (192.67,191.47) circle (  2.25);

\path[draw=drawColor,line width= 0.4pt,line join=round,line cap=round] (229.78,208.09) circle (  2.25);

\path[draw=drawColor,line width= 0.4pt,line join=round,line cap=round] (266.88,224.70) circle (  2.25);

\path[draw=drawColor,line width= 0.4pt,line join=round,line cap=round] (303.98,241.31) circle (  2.25);

\path[draw=drawColor,line width= 0.4pt,line join=round,line cap=round] (341.08,257.93) circle (  2.25);
\end{scope}
\begin{scope}
\path[clip] (  0.00,  0.00) rectangle (361.35,361.35);
\definecolor{drawColor}{RGB}{0,0,0}

\path[draw=drawColor,line width= 0.4pt,line join=round,line cap=round] ( 81.37, 30.00) -- (303.98, 30.00);

\path[draw=drawColor,line width= 0.4pt,line join=round,line cap=round] ( 81.37, 30.00) -- ( 81.37, 24.00);

\path[draw=drawColor,line width= 0.4pt,line join=round,line cap=round] (155.57, 30.00) -- (155.57, 24.00);

\path[draw=drawColor,line width= 0.4pt,line join=round,line cap=round] (229.78, 30.00) -- (229.78, 24.00);

\path[draw=drawColor,line width= 0.4pt,line join=round,line cap=round] (303.98, 30.00) -- (303.98, 24.00);

\node[text=drawColor,anchor=base,inner sep=0pt, outer sep=0pt, scale=  1.90] at ( 81.37,  8.40) {2};

\node[text=drawColor,anchor=base,inner sep=0pt, outer sep=0pt, scale=  1.90] at (155.57,  8.40) {4};

\node[text=drawColor,anchor=base,inner sep=0pt, outer sep=0pt, scale=  1.90] at (229.78,  8.40) {6};

\node[text=drawColor,anchor=base,inner sep=0pt, outer sep=0pt, scale=  1.90] at (303.98,  8.40) {8};

\path[draw=drawColor,line width= 0.4pt,line join=round,line cap=round] ( 32.40,108.41) -- ( 32.40,274.54);

\path[draw=drawColor,line width= 0.4pt,line join=round,line cap=round] ( 32.40,108.41) -- ( 26.40,108.41);

\path[draw=drawColor,line width= 0.4pt,line join=round,line cap=round] ( 32.40,191.47) -- ( 26.40,191.47);

\path[draw=drawColor,line width= 0.4pt,line join=round,line cap=round] ( 32.40,274.54) -- ( 26.40,274.54);

\node[text=drawColor,anchor=base east,inner sep=0pt, outer sep=0pt, scale=  1.90] at ( 20.40,101.87) {0};

\node[text=drawColor,anchor=base east,inner sep=0pt, outer sep=0pt, scale=  1.90] at ( 20.40,184.93) {5};

\node[text=drawColor,anchor=base east,inner sep=0pt, outer sep=0pt, scale=  1.90] at ( 20.40,268.00) {10};

\path[draw=drawColor,line width= 0.4pt,line join=round,line cap=round] ( 32.40, 30.00) --
	(352.95, 30.00) --
	(352.95,352.95) --
	( 32.40,352.95) --
	( 32.40, 30.00);
\end{scope}
\begin{scope}
\path[clip] ( 32.40, 30.00) rectangle (352.95,352.95);
\definecolor{drawColor}{RGB}{211,211,211}

\path[draw=drawColor,line width= 0.4pt,dash pattern=on 1pt off 3pt ,line join=round,line cap=round] ( 81.37, 30.00) -- ( 81.37,352.95);

\path[draw=drawColor,line width= 0.4pt,dash pattern=on 1pt off 3pt ,line join=round,line cap=round] (155.57, 30.00) -- (155.57,352.95);

\path[draw=drawColor,line width= 0.4pt,dash pattern=on 1pt off 3pt ,line join=round,line cap=round] (229.78, 30.00) -- (229.78,352.95);

\path[draw=drawColor,line width= 0.4pt,dash pattern=on 1pt off 3pt ,line join=round,line cap=round] (303.98, 30.00) -- (303.98,352.95);

\path[draw=drawColor,line width= 0.4pt,dash pattern=on 1pt off 3pt ,line join=round,line cap=round] ( 32.40,108.41) -- (352.95,108.41);

\path[draw=drawColor,line width= 0.4pt,dash pattern=on 1pt off 3pt ,line join=round,line cap=round] ( 32.40,191.47) -- (352.95,191.47);

\path[draw=drawColor,line width= 0.4pt,dash pattern=on 1pt off 3pt ,line join=round,line cap=round] ( 32.40,274.54) -- (352.95,274.54);
\definecolor{drawColor}{RGB}{0,0,0}

\path[draw=drawColor,line width= 1.2pt,line join=round,line cap=round] ( 44.27,108.41) --
	( 81.37,147.87) --
	(118.47,137.48) --
	(155.57,162.40) --
	(192.67,149.94) --
	(229.78,181.09) --
	(266.88,158.25) --
	(303.98,187.32) --
	(341.08,158.25);

\path[draw=drawColor,line width= 1.2pt,dash pattern=on 4pt off 4pt ,line join=round,line cap=round] ( 44.27,108.41) --
	( 81.37, 93.88) --
	(118.47, 82.60) --
	(155.57, 92.74) --
	(192.67, 77.53) --
	(229.78, 94.94) --
	(266.88, 71.40) --
	(303.98,102.21) --
	(341.08, 87.77);

\path[draw=drawColor,line width= 1.2pt,dash pattern=on 4pt off 4pt ,line join=round,line cap=round] ( 44.27,108.41) --
	( 81.37,201.86) --
	(118.47,192.37) --
	(155.57,232.06) --
	(192.67,222.36) --
	(229.78,267.25) --
	(266.88,245.10) --
	(303.98,272.44) --
	(341.08,228.73);
\definecolor{fillColor}{RGB}{255,255,255}

\path[draw=drawColor,line width= 0.4pt,line join=round,line cap=round,fill=fillColor,fill opacity=0.80] ( 32.40,352.95) rectangle (173.10,288.15);

\path[draw=drawColor,line width= 1.6pt,line join=round,line cap=round] ( 48.60,331.35) -- ( 81.00,331.35);

\path[draw=drawColor,line width= 1.6pt,dash pattern=on 4pt off 4pt ,line join=round,line cap=round] ( 48.60,309.75) -- ( 81.00,309.75);

\node[text=drawColor,anchor=base west,inner sep=0pt, outer sep=0pt, scale=  1.80] at ( 97.20,325.15) {$\bsmu$  \textcolor{white}{-}};

\node[text=drawColor,anchor=base west,inner sep=0pt, outer sep=0pt, scale=  1.80] at ( 97.20,303.55) {$\bsmu \pm \bssigma $  \textcolor{white}{-}};
\end{scope}
\end{tikzpicture}  }
        \caption{$\mu$ and $\sigma$ characteristics of \eqref{fig_toy_a}.}
        \label{fig_toy_d}
    \end{subfigure}
\begin{subfigure}[b]{0.325\textwidth}
\centering
\resizebox{.999\textwidth}{!}{ 
\begin{tikzpicture}[x=1pt,y=1pt]
\definecolor{fillColor}{RGB}{255,255,255}
\path[use as bounding box,fill=fillColor,fill opacity=0.00] (0,0) rectangle (361.35,361.35);
\begin{scope}
\path[clip] ( 32.40, 30.00) rectangle (352.95,352.95);
\definecolor{drawColor}{RGB}{255,255,255}

\path[draw=drawColor,line width= 0.4pt,line join=round,line cap=round] ( 44.27,125.02) circle (  2.25);

\path[draw=drawColor,line width= 0.4pt,line join=round,line cap=round] ( 81.37,141.64) circle (  2.25);

\path[draw=drawColor,line width= 0.4pt,line join=round,line cap=round] (118.47,158.25) circle (  2.25);

\path[draw=drawColor,line width= 0.4pt,line join=round,line cap=round] (155.57,174.86) circle (  2.25);

\path[draw=drawColor,line width= 0.4pt,line join=round,line cap=round] (192.67,191.47) circle (  2.25);

\path[draw=drawColor,line width= 0.4pt,line join=round,line cap=round] (229.78,208.09) circle (  2.25);

\path[draw=drawColor,line width= 0.4pt,line join=round,line cap=round] (266.88,224.70) circle (  2.25);

\path[draw=drawColor,line width= 0.4pt,line join=round,line cap=round] (303.98,241.31) circle (  2.25);

\path[draw=drawColor,line width= 0.4pt,line join=round,line cap=round] (341.08,257.93) circle (  2.25);
\end{scope}
\begin{scope}
\path[clip] (  0.00,  0.00) rectangle (361.35,361.35);
\definecolor{drawColor}{RGB}{0,0,0}

\path[draw=drawColor,line width= 0.4pt,line join=round,line cap=round] ( 81.37, 30.00) -- (303.98, 30.00);

\path[draw=drawColor,line width= 0.4pt,line join=round,line cap=round] ( 81.37, 30.00) -- ( 81.37, 24.00);

\path[draw=drawColor,line width= 0.4pt,line join=round,line cap=round] (155.57, 30.00) -- (155.57, 24.00);

\path[draw=drawColor,line width= 0.4pt,line join=round,line cap=round] (229.78, 30.00) -- (229.78, 24.00);

\path[draw=drawColor,line width= 0.4pt,line join=round,line cap=round] (303.98, 30.00) -- (303.98, 24.00);

\node[text=drawColor,anchor=base,inner sep=0pt, outer sep=0pt, scale=  1.90] at ( 81.37,  8.40) {2};

\node[text=drawColor,anchor=base,inner sep=0pt, outer sep=0pt, scale=  1.90] at (155.57,  8.40) {4};

\node[text=drawColor,anchor=base,inner sep=0pt, outer sep=0pt, scale=  1.90] at (229.78,  8.40) {6};

\node[text=drawColor,anchor=base,inner sep=0pt, outer sep=0pt, scale=  1.90] at (303.98,  8.40) {8};

\path[draw=drawColor,line width= 0.4pt,line join=round,line cap=round] ( 32.40,108.41) -- ( 32.40,274.54);

\path[draw=drawColor,line width= 0.4pt,line join=round,line cap=round] ( 32.40,108.41) -- ( 26.40,108.41);

\path[draw=drawColor,line width= 0.4pt,line join=round,line cap=round] ( 32.40,191.47) -- ( 26.40,191.47);

\path[draw=drawColor,line width= 0.4pt,line join=round,line cap=round] ( 32.40,274.54) -- ( 26.40,274.54);

\node[text=drawColor,anchor=base east,inner sep=0pt, outer sep=0pt, scale=  1.90] at ( 20.40,101.87) {0};

\node[text=drawColor,anchor=base east,inner sep=0pt, outer sep=0pt, scale=  1.90] at ( 20.40,184.93) {5};

\node[text=drawColor,anchor=base east,inner sep=0pt, outer sep=0pt, scale=  1.90] at ( 20.40,268.00) {10};

\path[draw=drawColor,line width= 0.4pt,line join=round,line cap=round] ( 32.40, 30.00) --
	(352.95, 30.00) --
	(352.95,352.95) --
	( 32.40,352.95) --
	( 32.40, 30.00);
\end{scope}
\begin{scope}
\path[clip] ( 32.40, 30.00) rectangle (352.95,352.95);
\definecolor{drawColor}{RGB}{211,211,211}

\path[draw=drawColor,line width= 0.4pt,dash pattern=on 1pt off 3pt ,line join=round,line cap=round] ( 81.37, 30.00) -- ( 81.37,352.95);

\path[draw=drawColor,line width= 0.4pt,dash pattern=on 1pt off 3pt ,line join=round,line cap=round] (155.57, 30.00) -- (155.57,352.95);

\path[draw=drawColor,line width= 0.4pt,dash pattern=on 1pt off 3pt ,line join=round,line cap=round] (229.78, 30.00) -- (229.78,352.95);

\path[draw=drawColor,line width= 0.4pt,dash pattern=on 1pt off 3pt ,line join=round,line cap=round] (303.98, 30.00) -- (303.98,352.95);

\path[draw=drawColor,line width= 0.4pt,dash pattern=on 1pt off 3pt ,line join=round,line cap=round] ( 32.40,108.41) -- (352.95,108.41);

\path[draw=drawColor,line width= 0.4pt,dash pattern=on 1pt off 3pt ,line join=round,line cap=round] ( 32.40,191.47) -- (352.95,191.47);

\path[draw=drawColor,line width= 0.4pt,dash pattern=on 1pt off 3pt ,line join=round,line cap=round] ( 32.40,274.54) -- (352.95,274.54);
\definecolor{drawColor}{RGB}{0,0,0}

\path[draw=drawColor,line width= 1.2pt,line join=round,line cap=round] ( 44.27,108.41) --
	( 81.37,137.53) --
	(118.47,122.85) --
	(155.57,144.47) --
	(192.67,123.42) --
	(229.78,151.40) --
	(266.88,123.98) --
	(303.98,145.03) --
	(341.08,123.42);

\path[draw=drawColor,line width= 0.8pt,dash pattern=on 4pt off 4pt ,line join=round,line cap=round] ( 44.27,108.41) --
	( 81.37, 84.81) --
	(118.47, 79.36) --
	(155.57, 75.66) --
	(192.67, 68.59) --
	(229.78, 66.47) --
	(266.88, 57.67) --
	(303.98, 65.75) --
	(341.08, 68.59);

\path[draw=drawColor,line width= 0.8pt,dash pattern=on 4pt off 4pt ,line join=round,line cap=round] ( 44.27,108.41) --
	( 81.37,190.25) --
	(118.47,166.34) --
	(155.57,213.27) --
	(192.67,178.25) --
	(229.78,236.33) --
	(266.88,190.29) --
	(303.98,224.31) --
	(341.08,178.25);
\definecolor{fillColor}{RGB}{255,255,255}

\path[draw=drawColor,line width= 0.4pt,line join=round,line cap=round,fill=fillColor,fill opacity=0.80] ( 32.40, 94.80) rectangle (201.00, 30.00);

\path[draw=drawColor,line width= 0.8pt,line join=round,line cap=round] ( 48.60, 73.20) -- ( 81.00, 73.20);

\path[draw=drawColor,line width= 0.8pt,dash pattern=on 4pt off 4pt ,line join=round,line cap=round] ( 48.60, 51.60) -- ( 81.00, 51.60);

\node[text=drawColor,anchor=base west,inner sep=0pt, outer sep=0pt, scale=  1.80] at ( 97.20, 67.00) {$\bsmu_{\text{E}}$  \textcolor{white}{-}};

\node[text=drawColor,anchor=base west,inner sep=0pt, outer sep=0pt, scale=  1.80] at ( 97.20, 45.40) {$\bsmu_{\text{E}} \pm \bssigma_{\text{E}} $  \textcolor{white}{-}};

\path[] ( 32.40,352.95) rectangle (318.84,266.55);

\node[text=drawColor,anchor=base west,inner sep=0pt, outer sep=0pt, scale=  1.80] at ( 97.20,325.15) {MAE($\bsmu-\bsmu_{\text{E}}$) = 1.41};

\node[text=drawColor,anchor=base west,inner sep=0pt, outer sep=0pt, scale=  1.80] at ( 97.20,303.55) {MAE($\bssigma-\bssigma_{\text{E}}$) = 0.5};

\node[text=drawColor,anchor=base west,inner sep=0pt, outer sep=0pt, scale=  1.80] at ( 97.20,281.95) {MAE($\Delta(\bsmu-\bsmu_{\text{E}})$) = 0.37};
\end{scope}
\end{tikzpicture}  }
        \caption{$\mu$ and $\sigma$ characteristics of \eqref{fig_toy_b}.}
        \label{fig_toy_e}
    \end{subfigure}
\begin{subfigure}[b]{0.325\textwidth}
\centering
\resizebox{.999\textwidth}{!}{ 
\begin{tikzpicture}[x=1pt,y=1pt]
\definecolor{fillColor}{RGB}{255,255,255}
\path[use as bounding box,fill=fillColor,fill opacity=0.00] (0,0) rectangle (361.35,361.35);
\begin{scope}
\path[clip] ( 32.40, 30.00) rectangle (352.95,352.95);
\definecolor{drawColor}{RGB}{255,255,255}

\path[draw=drawColor,line width= 0.4pt,line join=round,line cap=round] ( 44.27,125.02) circle (  2.25);

\path[draw=drawColor,line width= 0.4pt,line join=round,line cap=round] ( 81.37,141.64) circle (  2.25);

\path[draw=drawColor,line width= 0.4pt,line join=round,line cap=round] (118.47,158.25) circle (  2.25);

\path[draw=drawColor,line width= 0.4pt,line join=round,line cap=round] (155.57,174.86) circle (  2.25);

\path[draw=drawColor,line width= 0.4pt,line join=round,line cap=round] (192.67,191.47) circle (  2.25);

\path[draw=drawColor,line width= 0.4pt,line join=round,line cap=round] (229.78,208.09) circle (  2.25);

\path[draw=drawColor,line width= 0.4pt,line join=round,line cap=round] (266.88,224.70) circle (  2.25);

\path[draw=drawColor,line width= 0.4pt,line join=round,line cap=round] (303.98,241.31) circle (  2.25);

\path[draw=drawColor,line width= 0.4pt,line join=round,line cap=round] (341.08,257.93) circle (  2.25);
\end{scope}
\begin{scope}
\path[clip] (  0.00,  0.00) rectangle (361.35,361.35);
\definecolor{drawColor}{RGB}{0,0,0}

\path[draw=drawColor,line width= 0.4pt,line join=round,line cap=round] ( 81.37, 30.00) -- (303.98, 30.00);

\path[draw=drawColor,line width= 0.4pt,line join=round,line cap=round] ( 81.37, 30.00) -- ( 81.37, 24.00);

\path[draw=drawColor,line width= 0.4pt,line join=round,line cap=round] (155.57, 30.00) -- (155.57, 24.00);

\path[draw=drawColor,line width= 0.4pt,line join=round,line cap=round] (229.78, 30.00) -- (229.78, 24.00);

\path[draw=drawColor,line width= 0.4pt,line join=round,line cap=round] (303.98, 30.00) -- (303.98, 24.00);

\node[text=drawColor,anchor=base,inner sep=0pt, outer sep=0pt, scale=  1.90] at ( 81.37,  8.40) {2};

\node[text=drawColor,anchor=base,inner sep=0pt, outer sep=0pt, scale=  1.90] at (155.57,  8.40) {4};

\node[text=drawColor,anchor=base,inner sep=0pt, outer sep=0pt, scale=  1.90] at (229.78,  8.40) {6};

\node[text=drawColor,anchor=base,inner sep=0pt, outer sep=0pt, scale=  1.90] at (303.98,  8.40) {8};

\path[draw=drawColor,line width= 0.4pt,line join=round,line cap=round] ( 32.40,108.41) -- ( 32.40,274.54);

\path[draw=drawColor,line width= 0.4pt,line join=round,line cap=round] ( 32.40,108.41) -- ( 26.40,108.41);

\path[draw=drawColor,line width= 0.4pt,line join=round,line cap=round] ( 32.40,191.47) -- ( 26.40,191.47);

\path[draw=drawColor,line width= 0.4pt,line join=round,line cap=round] ( 32.40,274.54) -- ( 26.40,274.54);

\node[text=drawColor,anchor=base east,inner sep=0pt, outer sep=0pt, scale=  1.90] at ( 20.40,101.87) {0};

\node[text=drawColor,anchor=base east,inner sep=0pt, outer sep=0pt, scale=  1.90] at ( 20.40,184.93) {5};

\node[text=drawColor,anchor=base east,inner sep=0pt, outer sep=0pt, scale=  1.90] at ( 20.40,268.00) {10};

\path[draw=drawColor,line width= 0.4pt,line join=round,line cap=round] ( 32.40, 30.00) --
	(352.95, 30.00) --
	(352.95,352.95) --
	( 32.40,352.95) --
	( 32.40, 30.00);
\end{scope}
\begin{scope}
\path[clip] ( 32.40, 30.00) rectangle (352.95,352.95);
\definecolor{drawColor}{RGB}{211,211,211}

\path[draw=drawColor,line width= 0.4pt,dash pattern=on 1pt off 3pt ,line join=round,line cap=round] ( 81.37, 30.00) -- ( 81.37,352.95);

\path[draw=drawColor,line width= 0.4pt,dash pattern=on 1pt off 3pt ,line join=round,line cap=round] (155.57, 30.00) -- (155.57,352.95);

\path[draw=drawColor,line width= 0.4pt,dash pattern=on 1pt off 3pt ,line join=round,line cap=round] (229.78, 30.00) -- (229.78,352.95);

\path[draw=drawColor,line width= 0.4pt,dash pattern=on 1pt off 3pt ,line join=round,line cap=round] (303.98, 30.00) -- (303.98,352.95);

\path[draw=drawColor,line width= 0.4pt,dash pattern=on 1pt off 3pt ,line join=round,line cap=round] ( 32.40,108.41) -- (352.95,108.41);

\path[draw=drawColor,line width= 0.4pt,dash pattern=on 1pt off 3pt ,line join=round,line cap=round] ( 32.40,191.47) -- (352.95,191.47);

\path[draw=drawColor,line width= 0.4pt,dash pattern=on 1pt off 3pt ,line join=round,line cap=round] ( 32.40,274.54) -- (352.95,274.54);
\definecolor{drawColor}{RGB}{0,0,0}

\path[draw=drawColor,line width= 1.2pt,line join=round,line cap=round] ( 44.27,108.41) --
	( 81.37,113.60) --
	(118.47,129.18) --
	(155.57,118.79) --
	(192.67,139.56) --
	(229.78,123.99) --
	(266.88,143.71) --
	(303.98,129.18) --
	(341.08,139.56);

\path[draw=drawColor,line width= 0.8pt,dash pattern=on 4pt off 4pt ,line join=round,line cap=round] ( 44.27,108.41) --
	( 81.37,104.37) --
	(118.47, 92.26) --
	(155.57,100.34) --
	(192.67, 84.19) --
	(229.78, 96.30) --
	(266.88, 77.56) --
	(303.98, 92.26) --
	(341.08, 84.19);

\path[draw=drawColor,line width= 0.8pt,dash pattern=on 4pt off 4pt ,line join=round,line cap=round] ( 44.27,108.41) --
	( 81.37,122.83) --
	(118.47,166.09) --
	(155.57,137.25) --
	(192.67,194.93) --
	(229.78,151.67) --
	(266.88,209.87) --
	(303.98,166.09) --
	(341.08,194.93);
\definecolor{fillColor}{RGB}{255,255,255}

\path[draw=drawColor,line width= 0.4pt,line join=round,line cap=round,fill=fillColor,fill opacity=0.80] ( 32.40, 94.80) rectangle (210.65, 30.00);

\path[draw=drawColor,line width= 0.8pt,line join=round,line cap=round] ( 48.60, 73.20) -- ( 81.00, 73.20);

\path[draw=drawColor,line width= 0.8pt,dash pattern=on 4pt off 4pt ,line join=round,line cap=round] ( 48.60, 51.60) -- ( 81.00, 51.60);

\node[text=drawColor,anchor=base west,inner sep=0pt, outer sep=0pt, scale=  1.80] at ( 97.20, 67.00) {$\bsmu_{\text{W}}$  \textcolor{white}{-}};

\node[text=drawColor,anchor=base west,inner sep=0pt, outer sep=0pt, scale=  1.80] at ( 97.20, 45.40) {$\bsmu_{\text{W}} \pm \bssigma_{\text{W}} $  \textcolor{white}{-}};

\path[] ( 32.40,352.95) rectangle (323.66,266.55);

\node[text=drawColor,anchor=base west,inner sep=0pt, outer sep=0pt, scale=  1.80] at ( 97.20,325.15) {MAE($\bsmu-\bsmu_{\text{W}}$) =1.64};

\node[text=drawColor,anchor=base west,inner sep=0pt, outer sep=0pt, scale=  1.80] at ( 97.20,303.55) {MAE($\bssigma-\bssigma_{\text{W}}$) = 1.83};

\node[text=drawColor,anchor=base west,inner sep=0pt, outer sep=0pt, scale=  1.80] at ( 97.20,281.95) {MAE($\Delta(\bsmu-\bsmu_{\text{W}})$) = 2.27};
\end{scope}
\end{tikzpicture}  }
        \caption{$\mu$ and $\sigma$ characteristics of \eqref{fig_toy_c}.}
        \label{fig_toy_f}
    \end{subfigure}
    
%

    
\caption{Scenario reduction with 4 equally weighted scenarios which are reduced to 2 scenarios, containing the 
original setting (left), the reduction 
for the Energy distance (center) and Wasserstein distance (right). The top row  shows the trajectories with corresponding weights. The bottom row illustrates the mean $\bsmu=(\mu_1,\ldots,\mu_9)$ and the standard deviation $\bssigma = (\sigma_1,\ldots, \sigma_9)$ of the full scenario set and the two reductions.
 }
\label{fig_toy}
\end{figure}

 In this article, we present an alternative to the Wasserstein type distances based reduction techniques. We propose the energy distance (also known as energy statistic) for ensemble and scenario reduction, see \cite{szekely2013energy}. 
 The energy distance is a special case of the maximum mean discrepancy (MMD) which is popular in machine learning, see \cite{borgwardt2006integrating}.
 In applications, it often shows preferable statistical/stochastic properties for reduction problems in contrast to Wasserstein distance based approaches. To motivate this more appropriately, we show a small example, illustrated in Figure \ref{fig_toy}.

Fig \ref{fig_toy_a} shows the considered set of 4 scenarios $\bsy_1, \ldots, \bsy_4$ with $\bsy_i = (y_{i,1},\ldots,y_{i,9})$ on which we want to apply a scenario reduction so that the final set contains only 2 weighted trajectories. 
Figure \ref{fig_toy_d} illustrates the main characteristics of the original ensemble, the first two moments with respect to the time dimension: Formally, the 
 mean $\bsmu = \E \bsZ $ and standard deviation $\bssigma = \sqrt{\text{diag}(\cov \bsZ )}$  of the 9-dimensional uniform random variable $\bsZ$ on the four atoms $\bsy_1, \ldots, \bsy_4$, thus $\bsZ \sim \UU( \{\bsy_1, \bsy_2,\bsy_3,\bsy_4\}) $. Figures 
\ref{fig_toy_b} and \ref{fig_toy_c} show the trajectories of the best scenario reduction for the energy distance and Wasserstein distance with corresponding weights\footnote{The reduction is computed with respect to to $d_{\text{E},1}$ in \eqref{eq_energy_dist} and 
$d_{\text{W},1}$ in \eqref{eq_wasserstein} }. 
As in \ref{fig_toy_d}, Figures \ref{fig_toy_e} and \ref{fig_toy_f} provide the mean and standard deviation of the reduction results with some approximation characteristics.

To understand the reduction results, we study the considered scenario set. It contains 4 scenarios, the first three of them tend to increase over time, the remaining trajectory 4 (purple) tends to decrease.
Both scenario reduction techniques eliminate two of the increasing trajectories in the optimal solution. The Wasserstein reduction is based on the optimal redistribution rule (see \cite{dupavcova2003scenario}). In consequence, the weights of the eliminated trajectories are redistributed to the remaining paths. In Figure \ref{fig_toy_c}, the weights of trajectories 2 (green) and 3 (cyan) are redistributed to trajectory 1 (red) with a final weight of 75\%, while keeping the remaining 25\% for trajectory 4. On the first view this makes sense, as the trajectory 1 (red) tends to be the most centered one out of the three increasing trajectories. Indeed, if we would consider a classical transportation problem, this is the best choice. However, in many applications we use the remaining trajectories for sophisticated path-dependent optimization problems. Hence, we should care about the statistical/stochastic properties of the reduced scenario set. Characteristics like the mean and variance structure of the reduced scenario set should match the original data well. A method that transfers the full weight of a removed trajectory fully to another trajectory seems to be suboptimal. 
Moreover, in Figure \ref{fig_toy_a} we observe that 3 out of 4 trajectories exhibit a zig-zag pattern with positive peaks at even time steps. Thus, the overall mean $\bsmu$ in Figure \ref{fig_toy_d} shows the same behavior. Only trajectory 1 (red) deviates from this behavior and shows positive peaks at odd time steps. In the Wasserstein scenario reduction (see \ref{fig_toy_c}) this trajectory 1 (red) got a high weight of $75\%$. As consequence, the corresponding mean exhibits peaks at odd time steps as well. Thus, this time series characteristic is not covered at all, with likely huge negative consequences in decision making applications. 

In contrast, the scenario reduction based on the energy distance (see Figures \ref{fig_toy_b} and \ref{fig_toy_e}) better preserves distributional properties which are usually relevant for decision making. The energy distance method keeps the trajectories 2 (green) and 4 (purple) with non-trivial weight assignment. It notices to some extend that trajectory 1 (red) does not show similar path dependencies to the remaining trajectories and decides for the alternative trajectory 2 (green) as a representative of the three increasing paths. Thus, the original zig-zag pattern of the mean $\bsmu$ is preserved in the mean $\bsmu_{\text{E}}$ of the reduced scenario set. When comparing quantitative approximation measures, the mean absolute error (MAE) across $t$, of $\bsmu$, $\bssigma$ and $\Delta \bsmu = (\mu_1 -\mu_{0},\ldots, \mu_9 -\mu_{8}) $ (see Figures \ref{fig_toy_e} and \ref{fig_toy_f}) we notice that in all characteristics, the energy distance based scenario reduction is clearly favorable. Thus, it is worth to further investigate the energy distance for scenario reduction. 
We restrict ourself to reductions based on probability distances, more detailed on distances for multivariate distribution, sometimes referred to fan reductions.
Most importantly, we ignore nested distance aspects and scenario tree reduction.


%


In the next section, we introduce the considered setting for ensemble and scenario reduction.  
In Section 3, we discuss of the energy distance and its properties which we compare with the Wasserstein distance.
Afterwards we show applications to a Bernoulli random walk
which corresponds to a 
binary tree and real power system data. We close with a short discussion.

\section{Basics and notations}

Let $\YY = (\bsy_i)_{i \in \{1,\ldots, n\}} = ( \bsy_{1}, \ldots, \bsy_{n} )$ with $\bsy_i \in \R^k$ be a set of $n$ $k$-dimensional scenarios with associated weights/probabilities $\bsw =(w_1,\ldots, w_n)$ with $\bsone'\bsw =1$ and $w_i\geq 0$. 
We may regard  $(\YY,\bsw)$ as weighted ensemble with associated  random variable $\bsY = \bsY(\YY, \bsw)$ which has the (cumulative) distribution function
\begin{equation}
\bsF_{\bsY}(\bsz) = \sum_{i=1}^n w_i \mathbbm{1}\{\bsy_i \leq \bsz\}
\label{eq_cdf_Y} .
\end{equation}
In the equally weighted case $\bsw = \frac{1}{n}\bsone = (\frac{1}{n},\ldots, \frac{1}{n})'$, $\YY$ is usually referred to ensemble. 


In \emph{weighted ensemble reduction}, we are looking for a suitable subset $\bsc = \{c_1, \ldots, c_m\} \subseteq  \{1,\ldots,n\}$ with cardinality $\#(\bsc) = m\leq n$ and analyze 
the reduced subset $\XX = \XX(\bsc) = (\bsx_i)_{i \in \{1,\ldots, m\}} = (\bsy_i)_{i \in \bsc} = ( \bsy_{c_1}, \ldots, \bsy_{c_m} ) $ of $\YY$ with associated reduced weight vector $\bsv = \bsv(\bsc; \bsw)= \frac{1}{s}(w_i)_{i \in \bsc} = ( w_{c_1}, \ldots, w_{c_m} )/s $ with $s = \bsone'(w_i)_{i \in \bsc}$.
 Denote additionally by $\CC_m = \{ \bsc | \bsc \subseteq \{1,\ldots, n\}, \#(\bsc) = m\}$ the set of all subset of $\{1,\ldots, n\}$  with cardinality $m$. 
The target is now to find $\bsc \in \CC_m$ so that $\bsX = \bsX(\bsc) = \bsX(\bsc; \YY, \bsw)   \sim \bsF_{\XX}$ 
with distribution function  
\begin{equation}
\bsF_{\bsX}(\bsz) = \sum_{i=1}^m v_i \mathbbm{1}\{\bsx_i \leq \bsz\}
\label{eq_cdf_X} 
\end{equation}
is close to $\bsY \sim \bsF_{\YY}$ with respect to some distance, see 
Figure \ref{fig_intr_ens_red} and \ref{fig_intr_wens_red}. If $\bsw = \frac{1}{n}\bsone$ we get $\bsv = \frac{1}{m} \bsone$ and refer this special case as \emph{ensemble reduction} (see Fig. \ref{fig_intr_ens_red}). 

Now, let $d(\cdot,\cdot)$ be a distance between two $k$-dimensional random variables.
Examples are the Wasserstein and the energy distance which is discussed in more detail in the next section.
To recap, the given weighted set of $n$ scenarios $(\YY, \bsw)$ is going to be reduced to $(\XX, \bsv) = ( (\bsy_i)_{i \in \bsc}, (w_i)_{i \in \bsc})$ with $m\leq n$ scenarios corresponding to the subset $\bsc \subseteq \{1,\ldots,n\}$. 
Formally, this reduction problem can be written as a minimization:
\begin{equation}
 {\bsc}_{\text{opt}} (\YY,\bsw, m) = \argmin_{\bsc \in \CC_m} d(\bsX(\bsc; \YY,\bsw), \bsY(\YY,\bsw)) ,
 \label{eq_ensemble_reduction}
 \end{equation}
 with $\bsX(\bsc; \YY,\bsw) \sim \bsF_{\bsX}$ and $\bsY(\YY,\bsw) \sim \bsF_{\bsY}$ as defined 
 in \eqref{eq_cdf_X} and \eqref{eq_cdf_Y}.
 Then  $ ( (\bsy_i)_{i \in {\bsc}_{\text{opt}}}, (w_i)_{i \in {\bsc}_{\text{opt}}})$ is the optimally reduced ensemble with respect to $d$. In general, \eqref{eq_ensemble_reduction} is an NP-hard problem due to the combinatorial complexity. This means that we can only guarantee to find optimal solutions in low-dimensional cases.

\emph{Scenario reduction} is a generalization of the weighted ensemble reduction. We are looking to the optimal subset $\bsc \in \CC_m$ with  cardinality $m$ and for optimal weights $\bsa = (a_i)_{i \in \bsc} \in \R^m$. They are chosen so that the 
reweighted reduced ensemble $(\XX, \bsa)$ approximates well the 
original weighted scenario set $(\YY,\bsw)$ with respect to a distance $d$ (see Fig. \ref{fig_intr_scen_red}).
In general, the scenario reduction problem can be represented as by an inner continuous optimization for the weights $\bsa$ and an outer optimization for $\bsc$ which corresponds to an integer-valued optimization, see e.g. \cite{romisch2009scenario, rujeerapaiboon2018scenario}. 
Then an optimization procedure for a target reduction size $m$ can be written as 
\begin{enumerate}
 \item For all $\bsc \in \CC_m$ solve 
 \begin{equation}
 \bsa(\bsc) = \argmin_{\bsa\in [0,1]^m, \bsa'\bsone=1} d(\bsX(\bsc; \YY,\bsa), \bsY(\YY,\bsw))
 \label{eq_scen_red_inner_cont}
 \end{equation}
 with $\bsX(\bsc; \YY,\bsa) \sim \bsF_{\bsX}$ and $\bsY(\YY,\bsw) \sim \bsF_{\bsY}$ as defined 
 in \eqref{eq_cdf_X} and \eqref{eq_cdf_Y}.

 \item Compute 
\begin{equation}
\bsc_{\text{opt}} = \argmin_{\bsc \in \CC_m} d(\bsX(\bsc; \YY,\bsa(\bsc)), \bsY(\YY,\bsw))
 \label{eq_scen_red_outer_cont}
\end{equation}
and  return the optimal solution  $(\bsc_{\text{opt}}, \bsa_{\text{opt}}) = (\bsc_{\text{opt}}, \bsa(\bsc_{\text{opt}}))$
\end{enumerate}
As for ensemble reduction, $\CC_m$ is too large to solve the problem by brute force for the majority of practical problems.
Note that in general neither \eqref{eq_ensemble_reduction}, nor
 \eqref{eq_scen_red_inner_cont} and
 \eqref{eq_scen_red_outer_cont}
do attain a unique minimum. However, in many applications this is usually the case. Still, if the minimum is not unique, we might report and proceed with all minima or use another decision rule to report only one optimum.


\section{Why the energy distance for reduction problems?}



There is a wide range of plausible distances/metrics that measure the discrepancy between two random vectors or two multivariate distributions. 
Potential candidates are the disparity metric, total variation metric,
discrepancy metric, Hellinger distance and Wasserstein distances among others.
However, as mentioned in the introduction the Wasserstein metric is by far the most popular for scenario reduction. Hence, we recap its definition on $\R^k$:
\begin{equation}
 d_{\text{W}, p}(\bsX, \bsY) =  \left( \inf_{\gamma \in \Gamma(\bsX,\bsY) } \int_{\R^k \times \R^k} \|\bsx-\bsy\|_2^p d\, \gamma(\bsx,\bsy) \right)^{\frac{1}{p}} 
\label{eq_wasserstein} 
\end{equation}
where $\Gamma(\bsX,\bsY)$ is the set of all probability measures on $\R^k \times \R^k$ with the same marginals as $\bsX$ on the first $k$ coordinates and the same as $\bsY$ on the latter ones. 
$\Gamma(\bsX,\bsY)$ is also known as set of $\bsX$ and $\bsY$ couplings. 
As mentioned in the introduction, the popularity of the 
Wasserstein distance in scenario reduction is mainly due to  
available stability results for the Wasserstein distance in stochastic programming \cite{romisch2003stability} and the efficient reduction algorithm, see \cite{dupavcova2003scenario, romisch2009scenario}. The latter provides an efficient and explicit formula for the subproblem \eqref{eq_scen_red_inner_cont}  in scenario reduction. 
The resulting optimal redistribution rule (see e.g. \cite{growe2003scenario}) corresponds to a (mass) transportation problem. Here, it is important to remark that this algorithm does not require the explicit computation of the Wasserstein distance, which leads to a dramatic speed up compared to alternatives. 

Still, even for Wasserstein based scenario reduction, the NP-hard integer-valued optimization problem \eqref{eq_scen_red_outer_cont} is remaining. 
Thus, in ensemble reduction \eqref{eq_ensemble_reduction}
the advantages of the Wasserstein distance diminish as the inner optimization step in \eqref{eq_scen_red_inner_cont} is not required.  
 
The NP-hard integer-valued subproblem \eqref{eq_ensemble_reduction} or  \eqref{eq_scen_red_inner_cont} is usually solved by heuristics for higher dimensional problems. Typically, forward selection (FS), backward selection (FS) or similar algorithms like fast forward selection (FFS) or forward selection in wait-and-see clusters (FSWC) are applied, see e.g. \cite{growe2003scenario, romisch2009scenario, feng2013scenario}. However, other alternatives like particle swarm optimization, neural network bases on deep learning methods and other heuristics are applied as well, see \cite{li2019fast}. 

 The Wasserstein distance $d_{\text{W}, 1}$ with $p=1$ is often applied to scenario reduction as it satisfies 
 the duality theorem of Kantorovich and Rubinstein.
  Also note that in the $k=1$-dimensional case, the Wasserstein distance satisfies $d_{\text{W}, p}(X, Y) = \| F_{X}^{-1} - F_{Y}^{-1}\|_p$, see e.g. \cite{pflug2014multistage}. Thus,
  $ d_{\text{W}, 1}$ simplifies to
  \begin{equation}
    d_{\text{W}, 1}(X, Y)  = \int_\R | F_{X}(z)-F_{Y}(z) | d \, z ,
    \label{eq_wasserstein1d}
  \end{equation}
which is the $L^1$-distance between the two cumulative distribution functions.

%
A related discrepancy measure to   \eqref{eq_wasserstein1d} is the Cram\'er distance. It is given by
\begin{align}
 d_{C}(X, Y)  = 
 \int_\R | F_{X}(z)-F_{Y}(z) |^2 d \, z 
\label{eq_cramer_def}
\end{align}
 and measures the squared $L^2$-distance between the two cumulative distribution functions.  Thus, just from the intuitive point of view
$d_{\text{W}, 1}$ and $d_{C}$ should exhibit similar properties. 
The Cram\'er distance has a useful different representation given by
\begin{equation}
d_{C}(X,Y) = 
\E|X-Y| - \frac{1}{2}\E|X-X'| -\frac{1}{2} \E|Y-Y'|, 
\label{eq_cramer_repr}
\end{equation}
where $X'$ and $Y'$ are iid copies of $X$ and $Y$. 

 Now, the interesting part is that the representation \eqref{eq_cramer_repr}  allows for a natural generalization of the Cram\'er distance to $\R^k$, called energy distance, see \cite{szekely2013energy}. 
The \emph{energy distance} is defined as
\begin{equation}
d_{\text{E},p}(\bsX,\bsY) = 
\E \|\bsX-\bsY\|^p_2 - \frac{1}{2} \E \|\bsX-\bsX'\|^p_2 -\frac{1}{2} \E \|\bsY-\bsY'\|^p_2 
\label{eq_energy_dist}
\end{equation}
where $p\in (0,2)$ and again $\bsX'$ and $\bsY'$ are an iid copy of $\bsX$ and $\bsY$.
Obviously, for $k=1$ and $p=1$ \eqref{eq_energy_dist} corresponds to the Cram\'er distance. The case $p=1$ is regarded as the standard energy distance for $k>1$. Moreover, the energy distance is a special case of the maximum mean discrepancy (MMD), see e.g. \cite{borgwardt2006integrating}.

As pointed out in \cite{szekely2013energy, szekely2017energy}, the energy distance 
satisfies all axioms of a metric, especially that the 
energy distance is zero ($d_{\text{E},p}(\bsX,\bsY) = 0$) if and only if $\bsX$ and $\bsY$ share the same distribution (almost surely).
Moreover, it has useful transformation properties.
So $d_{\text{E},p}$ is scale equivariant, i.e.
for all constants $a\in \R$ there is a non-zero real-valued function $g$ such that it holds $d_{\text{E},p}(a\bsX, a\bsY) = g(a) d_{\text{E},p}(\bsX, \bsY)$. The energy distance satisfies the condition by choosing $g(z)=z^p$, which makes the choise $p=1$ to a natural candidate for applications. Furthermore, the energy distance exhibits rotational invariance,
i.e. for every orthonormal matrix $\bsA$ it holds $d_{\text{E},p}(\bsA\bsX, \bsA\bsY) = d_{\text{E},p}(\bsX, \bsY)$. It is interesting to note, that 
 the multivariate version of \eqref{eq_cramer_def} results in a measure that is not rotational invariant. This 
 is a hint that \eqref{eq_cramer_repr} is more suitable for a $k$-dimensional generalizations than \eqref{eq_cramer_def} which 
 focuses on the distances between the cumulative distribution functions. It also indicates that in higher dimensions the properties of the Wasserstein distance might deviate substantially from the energy distance. Moreover, we want to remark that also the Wasserstein distance is scale and rotational invariant. 
 It holds $d_{\text{W},p}(a\bsX, a\bsY) = |a| d_{\text{W},p}(\bsX, \bsY)$ and $d_{\text{W},p}(\bsA\bsX, \bsA\bsY) = d_{\text{W},p}(\bsX, \bsY)$ for all $p$, see \cite{muskulus2011wasserstein}. 

The probably most important property of the energy distance is an alternative representation by characteristic functions.
The energy distance can be rewritten as a weighted $L^2$-distance of the characteristic functions $\bsvarphi_{\bsX}$ and $\bsvarphi_{\bsY}$ 
with $\bsvarphi_{\bsX}(\bsz) = \E( e^{i\bsz'\bsX})$ and $\bsvarphi_{\bsY}(\bsz) = \E( e^{i\bsz'\bsY})$:
\begin{equation}
d_{\text{E},p}(\bsX,\bsY) = C_{k,p} \int_{\R^k } \frac{|\bsvarphi_{\bsX}(\bsz) - \bsvarphi_{\bsY}(\bsz)|^2 }{\|\bsz\|_2^{k+p}} \, d \bsz
\ \ \text{ with } \ \ C_{k,p} = \frac{\pi^{\frac{k}{2}} \Gamma(1-\frac{p}{2}) }{p2^{p}\Gamma(\frac{k+p}{2})}
\label{eq_energy_dist_characteristic}
\end{equation}
 Here, $C_{k,p}$ is a scaling constant which depends only on the dimension $k$ and $p$. 
 The weight function $\xi(\bsz) = \|\bsz\|_2^{-(k+p)}$ is a polynomial of the Euclidean norm depending on the dimension $k$. Interestingly, this 
 is the only choice of any continuous function $\xi$ such that 
  for some $C>0$
 the weighted $L^2$-distance 
$ C \int_{\R^k } \xi(\bsz) |\bsvarphi_{\bsX}(\bsz) - \bsvarphi_{\bsY}(\bsz)|^2 \, d \bsz $ 
 between the characteristic functions $\bsvarphi_{\bsX}$ and $\bsvarphi_{\bsY}$  is scale equivariant and rotational invariant.

The characterization by characteristic functions allows the energy distance to serve for powerful tests and the characterization of independence, see \cite{rizzo2016energy}.
For instance \cite{szekely2013energy} show that 
the energy distance based test for multivariate normality has a high testing power, higher than all commonly used alternatives.
Moreover, it can be utilized to test for non-parametric characteristics like symmetry or skewness. 
The other very crucial feature is that 
the energy distance allows the construction of the distance covariance and joint distance covariance, \cite{chakraborty2019distance}.
Those dependency measures can be used to construct tests for multivariate independence, which is remarkable on its own. According to \cite{chakraborty2019distance}, this is the only known scale and rotation invariant way to test for multivariate independence. As we want to maintain the dependency structure in the scenario paths properly the energy distance serves as a suitable candidate for reduction problems. 

In general, the considered distances for reduction problems have to be computed. This holds always for the energy distance $d_{\text{E},p}$ and in ensemble reduction also for the Wasserstein distance $d_{\text{W},p}$.
The Wasserstein distance can be computed by minimum matching and solving the linear programming \cite{nguyen2011wasserstein}
\begin{align}
d_{\text{W},p}(\bsX, \bsY) =& \min_{\bsq \in [0,1]^{\{1,\ldots,n\}\times \bsc}} 
\left( \sum_{(i,j)\in \{1,\ldots,n\}\times \bsc} q_{i,j}\|  \bsy_i - \bsy_j \|_2^p  \right)^{1/p}
\label{eq_wasserstein_dist_empiric}
\end{align}
such that $\sum_{i=1}^n q_{i,j} = w_j$ and $\sum_{j\in \bsc}^n q_{i,j} = w_i$. This problem can be solved using standard linear program methods. \cite{gottschlich2014shortlist} propose an alternative shortlist method for \eqref{eq_wasserstein_dist_empiric} 
which is based on the simplex algorithm and allows substantial speed improvements for multiple applications.
The energy distance $d_{\text{E},p}$ \eqref{eq_energy_dist}
can be calculated by equation:
\begin{align}
d_{\text{E},p}(\bsX, \bsY) =& 
\frac{1}{nm} \sum_{i=1}^n \sum_{j\in \bsc}  \| w_i \bsy_i - w_j \bsy_j \|_2^p \nonumber \\ 
&-
\frac{1}{2m^2} \sum_{i\in \bsc} \sum_{j\in \bsc}  \| w_i \bsy_i- w_j \bsy_j\|_2^p
-
\frac{1}{2n^2} \sum_{i=1}^n \sum_{i=1}^n  \| w_i \bsy_i- w_j \bsy_j\|_2^p .
\label{eq_energy_dist_empiric}
\end{align}
However, in reduction problems we do not have to compute equation \eqref{eq_energy_dist_empiric}, as the latter term remains unchanged for all $\bsc$.
Thus, it is sufficient to replace $d_{\text{E},p}$ by
\begin{align}
\text{ES}_{p}(\bsX, \bsY) = 
\frac{1}{nm} \sum_{i=1}^n \sum_{j\in \bsc}  \| w_i \bsy_i - w_j \bsy_j \|_2^p -
\frac{1}{2m^2} \sum_{k\in \bsc} \sum_{j\in \bsc}  \| w_j \bsy_j- w_k \bsy_k\|_2^p  
\label{eq_energy_score}
\end{align}
which is known as the \emph{energy score} which is popular in forecasting evaluation, see \cite{gneiting2007strictly}.
In forecast evaluation, usually one-sided measures are required. So one input is taken as random variable, the other input is treated as observation in $\R^k$ and as already materialized. Among forecasters the continuous ranked probability score (CRPS) is usually regarded as the suitable 1-dimensional forecasting criterion for distributions. The CRPS is the one-sided counterpart of the Cram\'er distance \eqref{eq_cramer_repr}. Concerning the computation of the energy score, we remark that the computational complexity for calculating the terms $\| w_i \bsy_i- w_j \bsy_j\|_2^p$ for all $i$ and $j$ is fast and of order $\OO(nm)$.

For scenario reduction, the Wasserstein based approaches have the speed advantage due to the mentioned explicit algorithm proposed by \cite{dupavcova2003scenario}. In these cases, the energy distance based method have a speed disadvantage. However, the inner optimization step in 
 \eqref{eq_scen_red_inner_cont} turns with \eqref{eq_energy_score} to
\begin{align}
 \bsa(\bsc) &= \argmin_{\bsa\in [0,1]^m, \bsa'\bsone=1} \text{ES}_p(\bsX(\bsc; \YY,\bsa), \bsY(\YY,\bsw))
) \\
&= \argmin_{\bsa\in [0,1]^m, \bsa'\bsone=1} \frac{1}{nm} \sum_{i=1}^n \sum_{j\in \bsc}  \| a_i \bsy_i - w_j \bsy_j\|_2^p -
\frac{1}{2m^2} \sum_{i\in \bsc} \sum_{j\in \bsc}  \| a_i \bsy_i - a_j \bsy_j\|_2^p ,
\label{eq_quadratic_prog_lc}
\end{align}
which is a quadratic program (QP) with linear constraints. 
There is plenty of literature available to solve quadratic optimizations, \cite{bao2011semidefinite}. In general, the matrix $\bsA= (a_{i,j})_{(i,j)\in{\bsc\times \bsc}}$ is indefinite. Thus, we are facing the most complicated QP which might be potentially NP-hard, see \cite{vavasis1992local}.

However, our empirical applications show that standard algorithms provide a solution in reasonable time. The reason is likely the very special structure of $\bsA$ which is a Euclidian distance matrix (EDM) and is mainly studied in the context of 
Euclidean  distance matrix completion problems (EDMCP). 
Such a matrix has full rank if $\bsy_i \neq \bsy_j$ for all $i,j\in\bsc$ with $i\neq j$, \cite{ball1992eigenvalues}.
Additionally,
\cite{hayden1993distance} provides a proof that $-\bsA$ has  exactly one negative eigenvalue.
As our feasible region of the QP is compact, we can apply quadratic programming theory (see \cite{vavasis1992local}). 
It follows that there exists a polynomial time algorithm for \eqref{eq_quadratic_prog_lc} that provides $\eps$-exact solutions. 
As $-\bsA$ has exactly one negative eigenvalue, this polynomial time is in fact quadratic. However, to our knowledge there is no algorithm for QPs with EDMs available so far where this property has been proven.
Still, it is fair to assume that any reasonable algorithm
available for solving \eqref{eq_quadratic_prog_lc} will benefit from the aforementioned structure of the problem.

\section{Applications}

In this section, we show different applications for ensemble and scenario reduction using the energy distance and the Wasserstein distance for comparison purpose.
First we study ensemble and scenario reduction for a Bernoulli random walk which corresponds to a binary tree. 
Afterwards, we consider scenario reduction for electricity load/demand/consumption and electricity prices\footnote{The considered data is available at https://transparency.entsoe.eu}, which are highly relevant in many energy systems applications. 
However, both considered real data studies are mainly for illustration purpose. We do not apply the reduced scenario sets in a subsequent optimization problem.

\subsection{Bernoulli random walk}

We consider a symmetric Bernoulli random walk (also called symmetric simple random walk and coin-tossing random walk), where $P[Y_t=Y_{t-1}-1]=P[Y_t=Y_{t-1}+1]=0.5$ with $Y_0=0$.
The resulting paths correspond to an equally weighted binary tree which branches every step by $-1$ and $1$. 
We consider $T=5$ time steps to keep the problem sufficiently small and allow for the computation of exact solutions. 
The resulting  $n=2^T=2^5=32$ paths in the scenario set are interpreted as a fan, not as a tree. 
Note that we ignore additional information due to the tree structure which could be included into the reduction technique by considering
nested distances, see e.g. \cite{pflug2014multistage}.

\begin{figure}
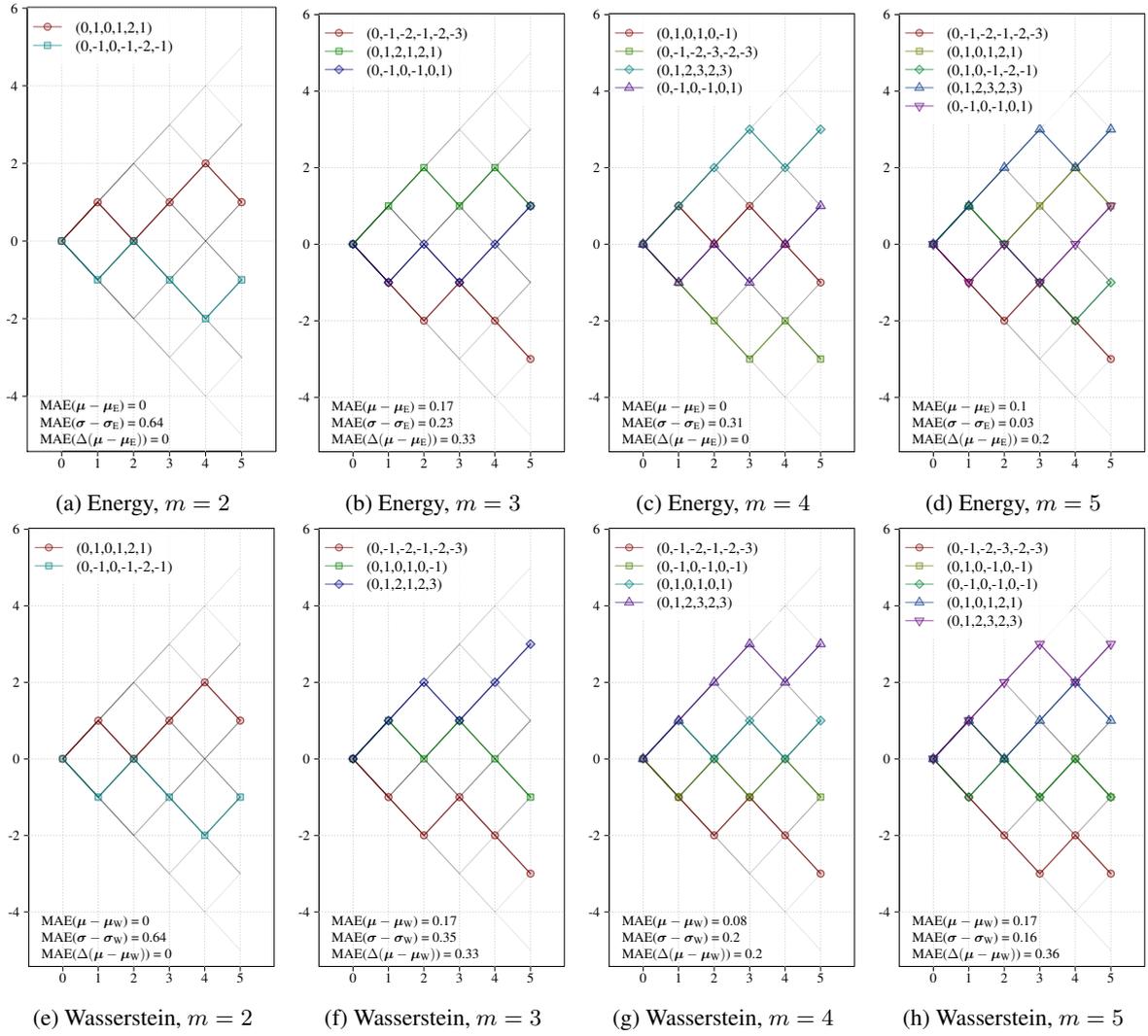


\begin{subfigure}[b]{0.24\textwidth}
\centering
\resizebox{.999\textwidth}{!}{\input{fig/binTree_H=5_cdn=2_energy.tex}}
        \caption{Energy, $m=2$}
        \label{fig_ens_energy_m_2}
    \end{subfigure}
\begin{subfigure}[b]{0.24\textwidth}
\centering
\resizebox{.999\textwidth}{!}{ \input{fig/binTree_H=5_cdn=3_energy.tex}  }
        \caption{Energy, $m=3$}
        \label{fig_ens_energy_m_3}
    \end{subfigure}
\begin{subfigure}[b]{0.24\textwidth}
\centering
\resizebox{.999\textwidth}{!}{ \input{fig/binTree_H=5_cdn=4_energy.tex}  }
        \caption{Energy, $m=4$}
        \label{fig_ens_energy_m_4}
    \end{subfigure}
\begin{subfigure}[b]{0.24\textwidth}
\centering
\resizebox{.999\textwidth}{!}{ \input{fig/binTree_H=5_cdn=5_energy.tex}  }
        \caption{Energy, $m=5$}
        \label{fig_ens_energy_m_5}
    \end{subfigure}

    \begin{subfigure}[b]{0.24\textwidth}
\centering
\resizebox{.999\textwidth}{!}{ \input{fig/binTree_H=5_cdn=2_wasserstein.tex}  }
        \caption{Wasserstein, $m=2$}
        \label{fig_ens_wasserstein_m_2}
    \end{subfigure}
\begin{subfigure}[b]{0.24\textwidth}
\centering
\resizebox{.999\textwidth}{!}{ \input{fig/binTree_H=5_cdn=3_wasserstein.tex}  }
        \caption{Wasserstein, $m=3$}
        \label{fig_ens_wasserstein_m_3}
    \end{subfigure}
\begin{subfigure}[b]{0.24\textwidth}
\centering
\resizebox{.999\textwidth}{!}{ \input{fig/binTree_H=5_cdn=4_wasserstein.tex}  }
        \caption{Wasserstein, $m=4$}
        \label{fig_ens_wasserstein_m_4}
    \end{subfigure}
\begin{subfigure}[b]{0.24\textwidth}
\centering
\resizebox{.999\textwidth}{!}{ \input{fig/binTree_H=5_cdn=5_wasserstein.tex}  }
        \caption{Wasserstein, $m=5$}
        \label{fig_ens_wasserstein_m_5}
    \end{subfigure}

\caption{Exact ensemble reduction for a symmetric Bernoulli random walk of size $T=5$ and $n = 2^T = 32$ scenarios.
The reduction solutions is performed for $m=2,\ldots,5$ scenarios (from left to right) for the energy distance $d_{\text{E},1}$ (top row) and the Wasserstein distance $d_{\text{W},1}$ (bottom row).
The resulting trajectories (top legends) the MAE of the approximation of $\mu_t$, $\sigma_t$ and $\Delta \mu_t$ are shown (bottom legends).}
\label{fig_ens_tree}
\end{figure}

\begin{figure}
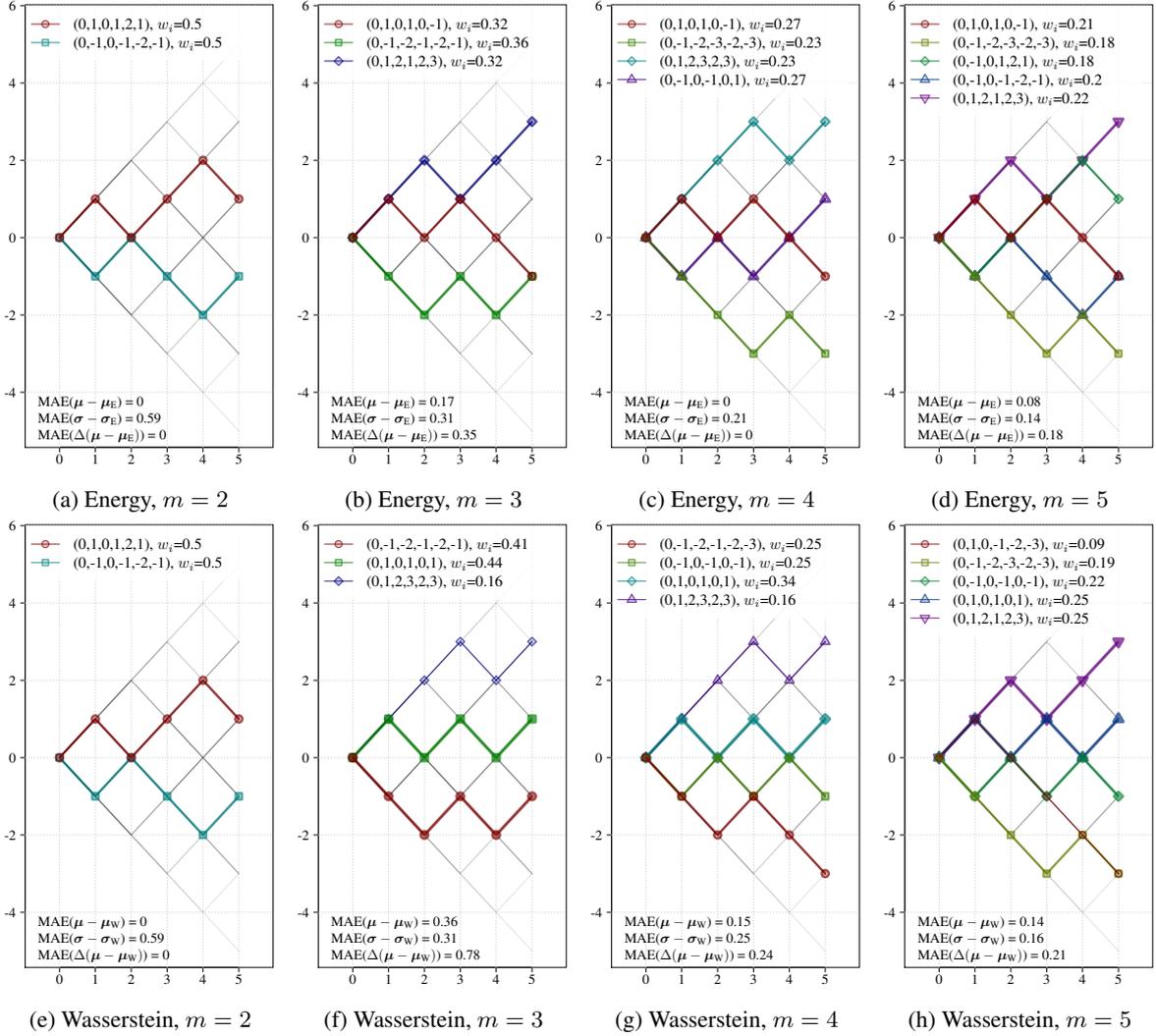


\begin{subfigure}[b]{0.24\textwidth}
\centering
\resizebox{.999\textwidth}{!}{ \input{fig/scenariobinTree_H=5_cdn=2_energy.tex}  }
        \caption{Energy, $m=2$}
        \label{fig_scen_energy_m_2}
    \end{subfigure}
\begin{subfigure}[b]{0.24\textwidth}
\centering
\resizebox{.999\textwidth}{!}{ \input{fig/scenariobinTree_H=5_cdn=3_energy.tex}  }
        \caption{Energy, $m=3$}
        \label{fig_scen_energy_m_3}
    \end{subfigure}
\begin{subfigure}[b]{0.24\textwidth}
\centering
\resizebox{.999\textwidth}{!}{ \input{fig/scenariobinTree_H=5_cdn=4_energy.tex}  }
        \caption{Energy, $m=4$}
        \label{fig_scen_energy_m_4}
    \end{subfigure}
\begin{subfigure}[b]{0.24\textwidth}
\centering
\resizebox{.999\textwidth}{!}{ \input{fig/scenariobinTree_H=5_cdn=5_energy.tex}  }
        \caption{Energy, $m=5$}
        \label{fig_scen_energy_m_5}
    \end{subfigure}

    \begin{subfigure}[b]{0.24\textwidth}
\centering
\resizebox{.999\textwidth}{!}{ \input{fig/scenariobinTree_H=5_cdn=2_wasserstein.tex}  }
        \caption{Wasserstein, $m=2$}
        \label{fig_scen_wasserstein_m_2}
    \end{subfigure}
\begin{subfigure}[b]{0.24\textwidth}
\centering
\resizebox{.999\textwidth}{!}{ \input{fig/scenariobinTree_H=5_cdn=3_wasserstein.tex}  }
        \caption{Wasserstein, $m=3$}
        \label{fig_scen_wasserstein_m_3}
    \end{subfigure}
\begin{subfigure}[b]{0.24\textwidth}
\centering
\resizebox{.999\textwidth}{!}{ \input{fig/scenariobinTree_H=5_cdn=4_wasserstein.tex}  }
        \caption{Wasserstein, $m=4$}
        \label{fig_scen_wasserstein_m_4}
    \end{subfigure}
\begin{subfigure}[b]{0.24\textwidth}
\centering
\resizebox{.999\textwidth}{!}{ \input{fig/scenariobinTree_H=5_cdn=5_wasserstein.tex}  }
        \caption{Wasserstein, $m=5$}
        \label{fig_scen_wasserstein_m_5}
    \end{subfigure}
\caption{Exact scenario reduction for a symmetric Bernoulli random walk of size $T=5$ and $n = 2^T = 32$ scenarios.
The reduction solutions is performed for $m=2,\ldots,5$ scenarios (from left to right) for the energy distance $d_{\text{E},1}$ (top row) and the Wasserstein distance $d_{\text{W},1}$ (bottom row).
The resulting trajectories and weights (top legends) the MAE of the approximation of $\mu_t$, $\sigma_t$ and $\Delta \mu_t$ are shown (bottom legends).}
\label{fig_scen_tree}
\end{figure}

We apply exact ensemble and scenario reduction to reduce the number of trajectories from $n=32$ to $m\in\{2,\ldots,5\}$ with respect to the energy distance $d_{\text{E},1}$ and the 
Wasserstein distance $d_{\text{W},1}$. Note that due to the symmetry  the resulting solutions are usually not unique. Thus, we report only one optimal solution.
Figures \ref{fig_ens_tree} and \ref{fig_scen_tree} show the results for the ensemble and scenario reduction problem.
They also provide the optimal trajectories and for the scenario reduction problem the resulting weights.
Similarly to the Figure \ref{fig_toy} in the introduction, we also report key characteristics for the approximation of the reduced scenario set to the original set of $n=32$ paths. In detail,  we report again the mean absolute error (MAE) of the approximation for the mean $\bsmu = (\mu_0,\ldots, \mu_5)$, standard deviation $\bssigma = 
(\sigma_0,\ldots, \sigma_5)$ and difference of the mean $\Delta \bsmu=(\mu_1 - \mu_0,\ldots, \mu_5 - \mu_4)$.

In Figures \ref{fig_ens_tree} and \ref{fig_scen_tree}, we observe that for $m=2$ both distances lead to the same optimal solution. However, for $m>2$ the solutions of the energy distance and the Wasserstein distance deviate for both the ensemble reduction problem and the scenario reduction problem. 
For ensemble reduction and $m=3$ (Fig. \ref{fig_ens_energy_m_3} and \ref{fig_ens_wasserstein_m_3}), we see that both solutions have the same MAE performance for the mean approximation but the energy distance approximates better the variance structure. For $m=4$ (Fig. \ref{fig_ens_energy_m_4} and \ref{fig_ens_wasserstein_m_4}), the results are mixed. The energy distance reduction maintains perfectly the mean behavior in contrast to the Wasserstein distance. But this comes at the cost of a worse explanation of the variance structure. For $m=5$ (Fig. \ref{fig_ens_energy_m_5} and \ref{fig_ens_wasserstein_m_5}),  the picture is clear. The energy distance has a better approximation in all considered characteristics. Another fact that can be observed for $m=5$ is that in the last step from $t=4$ to $t=5$,  the energy distance captures better the path dependency than the Wasserstein distance. In a binary tree, the ratio of branches which go upwards and downwards is equally distributed. As $m=5$ is odd this can never be perfectly captured in ensemble reduction. The energy distance reduction selects three increasing and two decreasing paths from $t=4$ to $t=5$. On the other hand, the optimal Wasserstein distance reduction provides one increasing trajectory and four decreasing trajectories which is an unnecessary bias from the true behavior.

For the scenario reduction (Fig. \ref{fig_scen_tree}) with $m>2$ the quantitative measures are even more in favor for the energy distance. Here, in all characteristics the energy distance reduction obtains preferable results. Moreover, we notice that the weights (Fig. \ref{fig_scen_energy_m_3}, \ref{fig_scen_energy_m_4} and \ref{fig_scen_energy_m_5}) of the energy distance are more equally distributed among all paths than for the Wasserstein distance (Fig. \ref{fig_scen_wasserstein_m_3}, \ref{fig_scen_wasserstein_m_4} and \ref{fig_scen_wasserstein_m_5}). This is not automatically an advantage. However, it indicates that the energy score potentially mixes the scenario properties in a stronger way as the weights are closer to the uniform distribution which corresponds to maximum entropy.

\subsection{Electricity load profiles}

We consider quarter-hourly German electricity demand data in two different scenario reduction settings. The first application is a small reduction problem where we can consider exact reduction methods. The second application uses forward selection for the scenario reduction and illustrates some reduction characteristics. In the first case, we apply scenario reduction using the energy distance $d_{\text{E},1}$ and the Wasserstein distance $d_{\text{W},1}$. In
the second example, we extend the set of considered distances 
to $d_{\text{E},p}$ with $p\in\{0.5,1,2\}$ and $d_{\text{W},p}$ with $p\in\{1,2,3\}$ to evaluate the impact of $p$.

\subsubsection{Exact reduction for demand profiles of 3rd Wednesday of each month.}

\begin{figure}
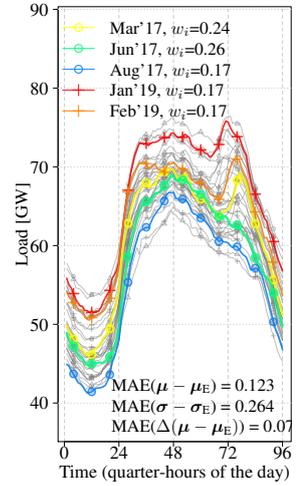
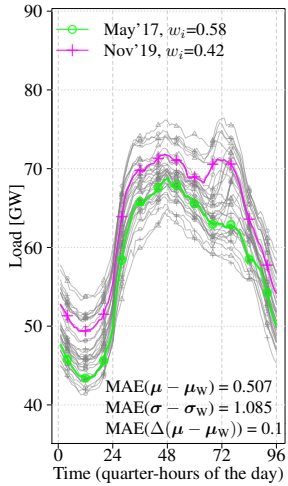
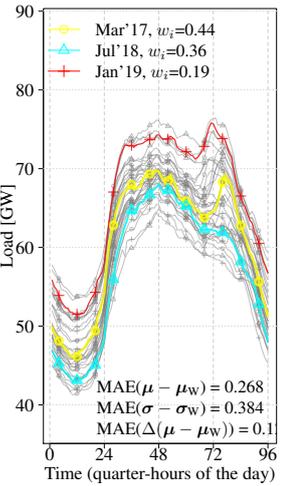
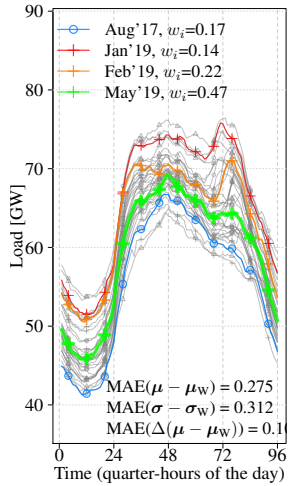

\begin{subfigure}[b]{0.48\textwidth}
\centering
\resizebox{.999\textwidth}{!}{\input{fig/load_exact_all.tex} }
        \caption{Considered scenarios}
        \label{fig_load_exact_all1}
    \end{subfigure}
    \begin{subfigure}[b]{0.48\textwidth}
\centering
\resizebox{.999\textwidth}{!}{\input{fig/load_exact_all2.tex} }
        \caption{Mean and Variance characteristics of \eqref{fig_load_exact_all1}}
        \label{fig_load_exact_all2}
    \end{subfigure}
    \begin{subfigure}[b]{0.24\textwidth}
\centering
\resizebox{.999\textwidth}{!}{\input{fig/load_exact_cdn=2_energy.tex} }
        \caption{Energy, $m=2$}
        \label{fig_load_exact_energy_m_2}
    \end{subfigure}
\begin{subfigure}[b]{0.24\textwidth}
\centering
\resizebox{.999\textwidth}{!}{\input{fig/load_exact_cdn=3_energy.tex} }
        \caption{Energy, $m=3$}
        \label{fig_load_exact_energy_m_3}
    \end{subfigure}
\begin{subfigure}[b]{0.24\textwidth}
\centering
\resizebox{.999\textwidth}{!}{\input{fig/load_exact_cdn=4_energy.tex} }
        \caption{Energy, $m=4$}
        \label{fig_load_exact_energy_m_4}
    \end{subfigure}
\begin{subfigure}[b]{0.24\textwidth}
\centering
\resizebox{.999\textwidth}{!}{\input{fig/load_exact_cdn=5_energy.tex} }
        \caption{Energy, $m=5$}
        \label{fig_load_exact_energy_m_5}
    \end{subfigure}
\begin{subfigure}[b]{0.24\textwidth}
\centering
\resizebox{.999\textwidth}{!}{\input{fig/load_exact_cdn=2_wasserstein.tex} }
        \caption{Wasserstein, $m=2$}
        \label{fig_load_exact_wasserstein_m_2}
    \end{subfigure}
    \begin{subfigure}[b]{0.24\textwidth}
\centering
\resizebox{.999\textwidth}{!}{\input{fig/load_exact_cdn=3_wasserstein.tex} }
        \caption{Wasserstein, $m=3$}
        \label{fig_load_exact_wasserstein_m_3}
    \end{subfigure}
\begin{subfigure}[b]{0.24\textwidth}
\centering
\resizebox{.999\textwidth}{!}{\input{fig/load_exact_cdn=4_wasserstein.tex} }
        \caption{Wasserstein, $m=4$}
        \label{fig_load_exact_wasserstein_m_4}
    \end{subfigure}
\begin{subfigure}[b]{0.24\textwidth}
\centering
\resizebox{.999\textwidth}{!}{\input{fig/load_exact_cdn=5_wasserstein.tex} }
        \caption{Wasserstein, $m=5$}
        \label{fig_load_exact_wasserstein_m_5}
    \end{subfigure}
\caption{Exact scenario reduction for demand profiles with $n = 36$ scenarios.
The top row shows the $n=36$ scenarios (top left) with expected value and standard deviation characteristics (top right).
The reduction solutions is performed for $m=2,\ldots,5$ scenarios (from left to right, center and bottom) for the energy distance $d_{\text{E},1}$ (center row) and the Wasserstein distance $d_{\text{W},1}$ (bottom row).
The resulting trajectories and weights (top legends) the MAE of the approximation of $\mu_t$, $\sigma_t$ and $\Delta \mu_t$ are shown (bottom legends).}
\label{fig_exact_load}
\end{figure}

This example considers electricity load data from January 2017 to December 2019. For each month the load profile of every 3rd Wednesday of each month is considered as the scenario set. 
This gives us $n=3\times12=36$ historic demand trajectories in total. Thus, exact scenario reduction is computationally feasible, at least for some settings with small $m$. The consideration of every third 3rd Wednesday in month is sometimes considered in power systems literature as well, see e.g. \cite{ andrychowicz2017review}.
The considered data and overall reduction results for $m=2,\ldots,5$ are given in Figure \ref{fig_exact_load}. 
Figures       \ref{fig_load_exact_all1} and         \ref{fig_load_exact_all2} depict the 36 path for the scenario reduction problem and the mean and variance characteristics.
Figure \ref{fig_load_exact_energy_m_2} to Figure \ref{fig_load_exact_wasserstein_m_5} show the results of exact scenario reduction for the energy distance $d_{\text{E},1}$ and Wasserstein distance $d_{\text{E},1}$ with the same MAE output measure as used before.

We observe that the overall behavior is similarly to Bernoulli random walk example but less clear in favor for the energy distance. 
The energy distance reduction results tend to have a superior behavior concerning the mean properties. Concerning the approximation of the correct variance structure, the results are mixed. For $m=2$ and $m=3$, the Wasserstein distance covers the true variance pattern slightly better than the energy distance reduction. Vice versa for $m=4$ and $m=5$, where the energy distance has preferable variance behavior. 
Again, we see that the weights of the energy distance reduction are more equally distributed. For instance for $m=4$ (Figures \ref{fig_load_exact_energy_m_4} and \ref{fig_load_exact_wasserstein_m_4}), the Wasserstein distance has a maximum weight of 0.47 for the May'19 trajectory and a minimum weight of 0.14 for the Jan'19 trajectory. In contrast, the energy distance reduction leads to a maximum and minimum weight of 0.30 and 0.21. Thus, we expect more stable and better mixed results from energy distance reductions. Further, we observe that the Wasserstein distance reduction has the tendency to choose slightly more extreme trajectories. Consider e.g. $m=4$: The Wasserstein distance selected Aug'17 as representative summer trajectory. However, this Aug'17 trajectory has the lowest night load among all considered trajectories in the scenario set. The energy distance selects the Jul'19 path for the representative summer trajectory. This has a slightly higher overall demand level, also during the night hours.
For $m=3$, we observe a similar selection pattern. The energy distance chooses Dec'18 as representative winter trajectory whereas the Wasserstein distance selects the Jan'19. The latter has a more distinct evening peak.

\subsubsection{Forward reduction for demand profiles.}

In this larger scenario reduction, application we consider all daily electricity load profiles from 2019 as set of original trajectories. The $n=365$ trajectories are reduced to $m=2,\ldots,100$ scenarios. As exact reduction is computationally not feasible, we consider forward selection, see e.g. \cite{heitsch2007note}. We use simple 1-step forward selection where we add in every iteration one trajectory to the scenario set. This, allows us to compare properties of the reduced set of trajectories with respect to the energy and Wasserstein distance, where we first focus on $d_{\text{E},1}$ and $d_{\text{W},1}$.

\begin{figure}
\begin{subfigure}[b]{0.48\textwidth}
\centering
\resizebox{.999\textwidth}{!}{\input{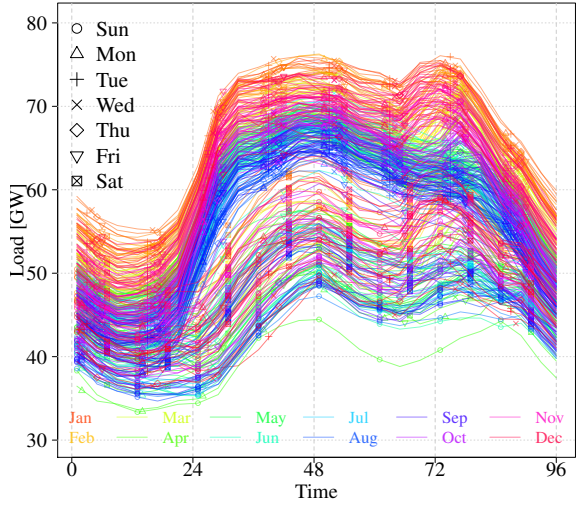} }
        \caption{Considered scenarios}
        \label{load_forw_all}
    \end{subfigure}
    \begin{subfigure}[b]{0.48\textwidth}
\centering
\resizebox{.999\textwidth}{!}{\input{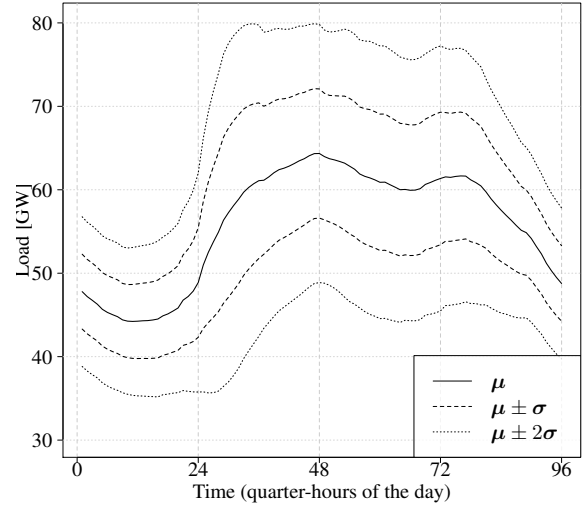} }
        \caption{Mean and Variance characteristics of \eqref{load_forw_all} }
        \label{load_forw_allb}
    \end{subfigure}
    \begin{subfigure}[b]{0.48\textwidth}
\centering
\resizebox{.999\textwidth}{!}{\input{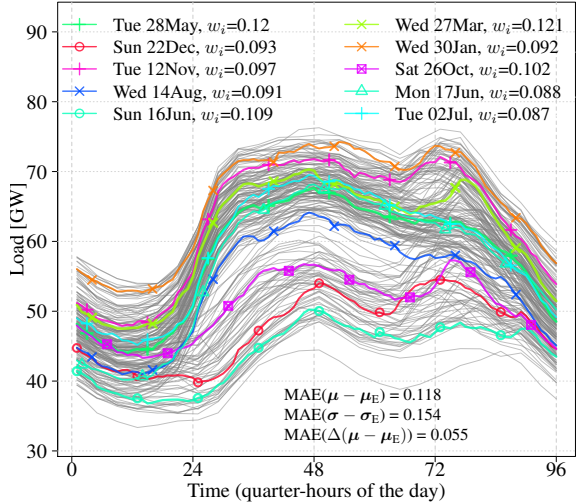} }
        \caption{Reduction using the energy distance}
        \label{fig_forw_10energy}
    \end{subfigure}
    \begin{subfigure}[b]{0.48\textwidth}
\centering
\resizebox{.999\textwidth}{!}{\input{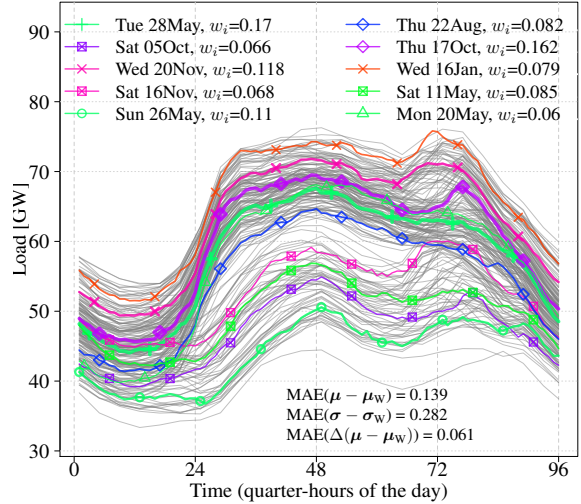} }
        \caption{Reduction using the Wasserstein distance}
        \label{fig_forw_10wasserstein}
    \end{subfigure}
\caption{Scenario reduction using forward selection for demand profiles with $n = 365$ scenarios.
The top row shows the $n=365$ scenarios (top left) with expected value and standard deviation characteristics (top right).
The reduction solutions for $m=10$ scenarios for the energy distance $d_{\text{E},1}$ (center left) and the Wasserstein distance $d_{\text{W},1}$ (center right) with characteristics.}
\label{fig_forw_load}
\end{figure}

\begin{figure}
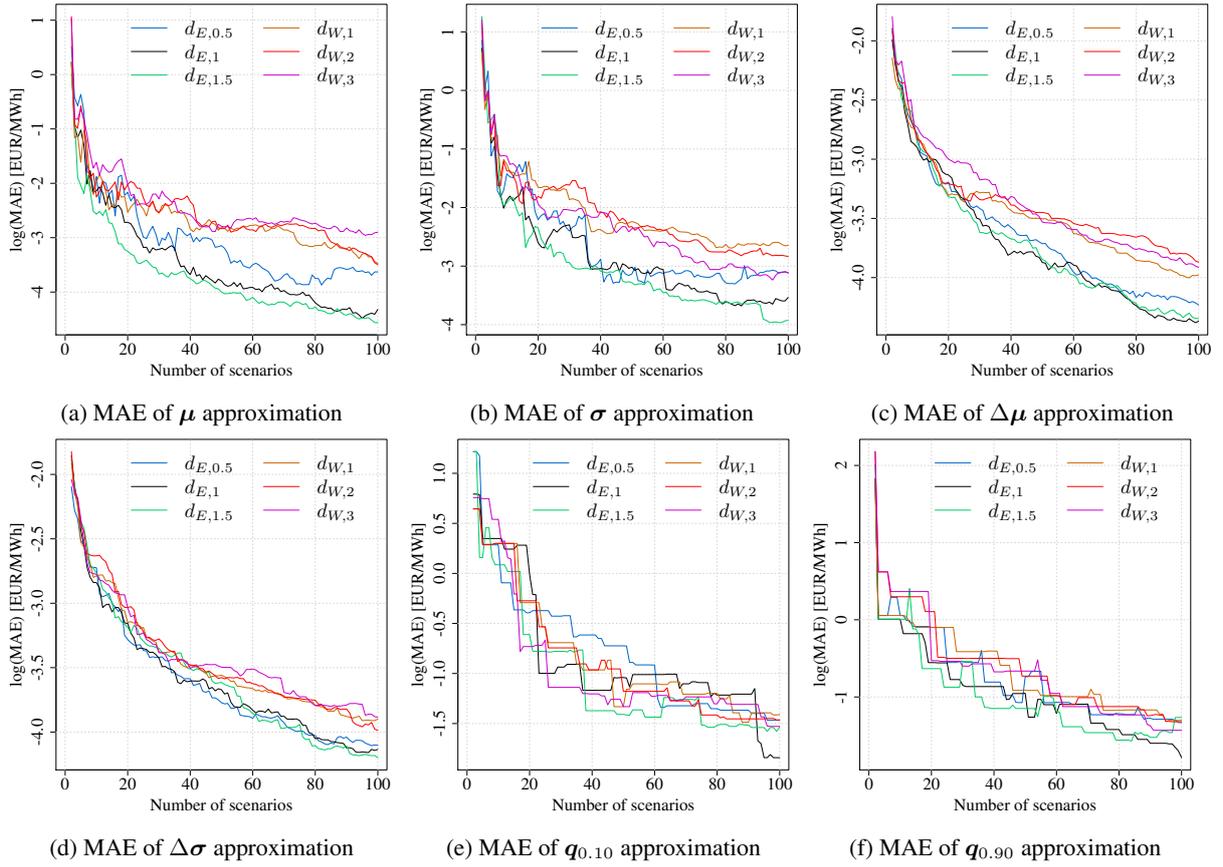

        \begin{subfigure}[b]{0.325\textwidth}
\centering
\resizebox{.999\textwidth}{!}{\input{fig/load_forw_cdn=100_mu.tex} }
        \caption{MAE of $\bsmu$ approximation}
        \label{fig_forw_mu}
    \end{subfigure}
        \begin{subfigure}[b]{0.325\textwidth}
\centering
\resizebox{.999\textwidth}{!}{\input{fig/load_forw_cdn=100_sd.tex} }
        \caption{MAE of $ \bssigma$ approximation}
        \label{fig_forw_sd}
    \end{subfigure}
        \begin{subfigure}[b]{0.325\textwidth}
\centering
\resizebox{.999\textwidth}{!}{\input{fig/load_forw_cdn=100_dmu.tex} }
        \caption{MAE of $\Delta \bsmu$ approximation}
        \label{fig_forw_dmu}
    \end{subfigure}
        \begin{subfigure}[b]{0.325\textwidth}
\centering
\resizebox{.999\textwidth}{!}{\input{fig/load_forw_cdn=100_dsd.tex} }
        \caption{MAE of $\Delta \bssigma$ approximation}
        \label{fig_forw_dsd}
    \end{subfigure}
        \begin{subfigure}[b]{0.325\textwidth}
\centering
\resizebox{.999\textwidth}{!}{\input{fig/load_forw_cdn=100_q10.tex} }
        \caption{MAE of $\bsq_{0.10}$ approximation}
        \label{fig_forw_q10}
    \end{subfigure}
        \begin{subfigure}[b]{0.325\textwidth}
\centering
\resizebox{.999\textwidth}{!}{\input{fig/load_forw_cdn=100_q90.tex} }
        \caption{MAE of $\bsq_{0.90}$ approximation}
        \label{fig_forw_q90}
    \end{subfigure}
\caption{Scenario reduction results by the energy distances ($d_{\text{E},0.5}$, $d_{\text{E},1}$, $d_{\text{E},1.5}$) and Wasserstein distances ($d_{\text{W},1}$, $d_{\text{W},2}$,$d_{\text{W},3}$).
 It shows the approximation accuracy in logarithm of the MAE (mean absolute error) of $\bsmu$ (top left), $\bssigma$ (top center), $\Delta \bsmu$ (top right),
$\Delta \bssigma$ (bottom left), 10\%-quantile $\bsq_{0.10}$ (bottom center) and 90\%-quantile $\bsq_{0.90}$ (bottom right) for $m=2,\ldots,100$ scenarios of the $n=365$ scenarios.}
\label{fig_forw_load_MAE}
\end{figure}

Figure \ref{fig_forw_load} presents the setting and the results of the  
reduction task. The top row (Figures \ref{load_forw_all} and  \ref{load_forw_allb}) illustrate the original set of 365 trajectories with mean and variance characteristics. Figures \ref{fig_forw_10energy} and \ref{fig_forw_10wasserstein} depict the scenario reduction results for the energy and Wasserstein distance for $m=10$ final scenarios. They also include the final weights and MAE approximation errors of $\mu_t$, $\sigma_t$ and $\Delta \mu_t$.
Additionally, Figures \eqref{fig_forw_mu}, \eqref{fig_forw_sd} and \eqref{fig_forw_dmu} depict the three MAE errors for all considered scenario reduction sizes up to $m=100$. Due to the high variations of the MAEs with respect to $m$, we added a monotonic smoothing spline fit to allow for better interpretation.

First, we interpret the results for $m=10$ (see \ref{fig_forw_10energy} and \ref{fig_forw_10wasserstein}). We observe that the results are similar to the exact reduction results we have seen before. The energy distance seems to provide more equally distributed weights for each trajectory than the Wasserstein distance. The weights vary between 0.087 and 0.121 whereas the Wasserstein distance weights vary between 0.060 and 0.170. Thus, the spread is more than 3 times as high. Moreover, we observe that both methods provide 10 trajectories that cover the general behavior of the data. They include working days and weekend days, such as days in summer and winter. Still, again the energy distance seems to spread the annual characteristic in a better way. 
As we apply forward selection we might also interpret the order of the considered trajectories. We see that both algorithm choose in the first step the same trajectory from Tue 28May but deviate from the second step onwards. A plausible result is that both algorithms choose a weekend day as second trajectory. Concerning the overall weekend representation, the energy distance reduction chooses a winter Sunday, a summer Sunday and a fall Saturday whereas the Wasserstein based reduction chooses two fall Saturdays, a spring Saturday and a spring Sunday. Again, the energy distance results seem to be more dispersed.
The provided MAE values draw the same picture: The energy distance provides better mean and variance approximations.

Now we turn towards the results of the MAE measures for different reduction sizes $m$. They are provided in Figure \ref{fig_forw_load_MAE} for $d_{\text{E},0.5}$, $d_{\text{E},1}$, 
$d_{\text{E},1.5}$, $d_{\text{W},1}$, $d_{\text{W},2}$ and $d_{\text{W},3}$.
 Next to the mean approximation ($\bsmu = (\mu_1,\ldots,\mu_{96})$), the standard deviation $\bssigma= (\sigma_1,\ldots,\sigma_{96})$ and the mean difference ($\Delta \bsmu = (\mu_2-\mu_1,\ldots,\mu_{96}-\mu_{95})$) we consider also
the standard deviation difference ($\Delta \bssigma = (\sigma_2-\sigma_1,\ldots,\sigma_{96}-\sigma_{95})$) and the 10\%- and 90\%-quantile processes ($ \bsq_{0.10} = (q_{0.10,1},\ldots,q_{0.10,96})$ and $\bsq_{0.90} = (q_{0.90,1},\ldots,q_{0.90,96})$). 
%
First, we see that all approximations tend to get better when the number of scenarios increases.
All energy distances show better mean and variance statistics (see 
\ref{fig_forw_mu}, \ref{fig_forw_sd}, \ref{fig_forw_dmu}, \ref{fig_forw_dsd}) than the Wasserstein distances.
For the 10\% and 90\% quantiles (see 
\ref{fig_forw_q10}, \ref{fig_forw_q90}), this is not the case. Here, no distance is clearly preferable.
Among the considered energy distances, $d_{\text{E},p}$ tends to 
have slightly better results for the mean and variance approximations 
for larger $p$. 
%
%

\subsection{Scenario reduction for electricity prices.}

In this final example we want to show an illustration for a larger real data set that is more heavy tailed and thus sensitive to outliers. Therefore, we consider 10 years of recent hourly German day-ahead electricity spot price profiles, from 01.07.2010 to 30.06.2020.
Table \ref{tab_summary} shows some summary statistics of the considered data set and their first difference.
We observe that the data set contains some positive and negative 
price spikes indicated by the large (in absolute terms) minimum and maximum, characteristic for electricity price data, see
\cite{ziel2018probabilistic}.
We reduce the $n=3653$ daily profile trajectories $\bsy_i = (y_{i,1},\ldots, y_{i,24})$ with $i\in \{1,\ldots, n\}$ for 
the energy and Wasserstein distance by evaluating additionally the impact of the parameter $p$ in both definitions by evaluating again 
$d_{\text{E},0.5}$, $d_{\text{E},1}$, $d_{\text{E},1.5}$, 
$d_{\text{W},1}$, $d_{\text{W},2}$ and $d_{\text{W},3}$.
As for the load profile, we consider a 1-step foreward selection as reduction algorithm. 

\begin{table}[ht]
\centering
\resizebox{.99\textwidth}{!}{
\begin{tabular}{r|rrrrrrrrrrrr}
  \hline
 & $\mu$ & $\sigma$ & min & max & $q_{0.10}$ & $q_{0.25}$ & $q_{0.50}$ & $q_{0.75}$ & $q_{0.90}$ & $\rho$ & $\gamma$ & $\kappa$ \\ 
  \hline
$\bsy_i $ & 37.68 & 17.08 & -221.99 & 210.00 & 18.77 & 27.96 & 36.93 & 47.95 & 58.43 & 0.74 & -0.51 & 11.76 \\ 
  $\Delta \bsy_i $ & 0.11 & 5.64 & -102.76 & 156.94 & -5.28 & -2.36 & -0.20 & 2.20 & 6.20 & -0.03 & 0.92 & 33.40 \\ 
   \hline
\end{tabular}
}
\caption{Summary statistics of electricity price data $\bsy_1,\ldots, \bsy_{3653}$:
mean ($\mu$), standard deviation ($\sigma$), minimum (min), maximum (max), quantiles for $10\%$, $25\%$, $50\%$, $75\%$, $90\%$ ($q_{0.10}$, $q_{0.25}$, $q_{0.50}$, $q_{0.75}$, $q_{0.90}$), 
mean correlation ($\rho$), skewness ($\gamma$) and kurtosis ($\kappa$) across all $n=3653$ trajectories.
}
\label{tab_summary}
\end{table}

We conduct the scenario reduction for a scenario size of $m=2,\ldots,100$ and evaluate the approximation accuracy of specific characteristics. We consider as before the mean ($\bsmu = (\mu_1,\ldots,\mu_{24})$), the standard deviation $\bssigma= (\sigma_1,\ldots,\sigma_{24})$,  difference in $\bsmu$ and $\bssigma$ ($\Delta \bsmu = (\mu_2-\mu_1,\ldots,\mu_{24}-\mu_{23})$, $\Delta \bssigma = (\sigma_2-\sigma_1,\ldots,\sigma_{24}-\sigma_{23})$), and the 10\%- and 90\%-quantile processes ($ \bsq_{0.10} = (q_{0.10,1},\ldots,q_{0.10,24})$ and $\bsq_{0.90} = (q_{0.90,1},\ldots,q_{0.90,24})$).
Figure \ref{fig_forw_price} depicts the results of the scenario reduction for the six considered distances. 
In general, we observe that again the energy distance seems to be superior to the Wasserstein distance in the mean related measures (see \ref{fig_forwp_mu} and \ref{fig_forwp_dmu}). 
For the energy distance $d_{\text{E},p}$, the approximation results tend to get worse with decreasing $p$. For the Wasserstein distance $d_{\text{W},p}$, the mean and variance characteristics also improve with increasing $p$. However, for the considered quantiles the picture is rather the other way around.

\begin{figure}
        \begin{subfigure}[b]{0.325\textwidth}
\centering
\resizebox{.999\textwidth}{!}{\input{fig/price2_forw_cdn=100_mu.tex} }
        \caption{MAE of $\bsmu$ approximation}
        \label{fig_forwp_mu}
    \end{subfigure}
        \begin{subfigure}[b]{0.325\textwidth}
\centering
\resizebox{.999\textwidth}{!}{\input{fig/price2_forw_cdn=100_sd.tex} }
        \caption{MAE of $ \bssigma$ approximation}
        \label{fig_forwp_sd}
    \end{subfigure}
        \begin{subfigure}[b]{0.325\textwidth}
\centering
\resizebox{.999\textwidth}{!}{\input{fig/price2_forw_cdn=100_dmu.tex} }
        \caption{MAE of $\Delta \bsmu$ approximation}
        \label{fig_forwp_dmu}
    \end{subfigure}
        \begin{subfigure}[b]{0.325\textwidth}
\centering
\resizebox{.999\textwidth}{!}{\input{fig/price2_forw_cdn=100_dsd.tex} }
        \caption{MAE of $\Delta \bssigma$ approximation}
        \label{fig_forwp_dsd}
    \end{subfigure}
        \begin{subfigure}[b]{0.325\textwidth}
\centering
\resizebox{.999\textwidth}{!}{\input{fig/price2_forw_cdn=100_q10.tex} }
        \caption{MAE of $\bsq_{0.10}$ approximation}
        \label{fig_forwp_q10}
    \end{subfigure}
        \begin{subfigure}[b]{0.325\textwidth}
\centering
\resizebox{.999\textwidth}{!}{\input{fig/price2_forw_cdn=100_q90.tex} }
        \caption{MAE of $\bsq_{0.90}$ approximation}
        \label{fig_forwp_q90}
    \end{subfigure}
\caption{Scenario reduction results by energy and Wasserstein distances ($d_{\text{E},0.5}$, $d_{\text{E},1}$, $d_{\text{E},1.5}$, $d_{\text{W},1}$, $d_{\text{W},2}$, $d_{\text{W},3}$).
 It shows the approximation accuracy in MAE (mean absolute error) of $\bsmu$ (top left), $\bssigma$ (top center), $\Delta \bsmu$ (top right),
$\Delta \bssigma$ (bottom left), 10\%-quantile $\bsq_{0.10}$ (bottom center) and 90\%-quantile $\bsq_{0.90}$ (bottom right) for $m=2,\ldots,100$ scenarios of the $n=3653$ scenarios with monotonic smoothing spline fit.}
\label{fig_forw_price}
\end{figure}

\begin{figure}
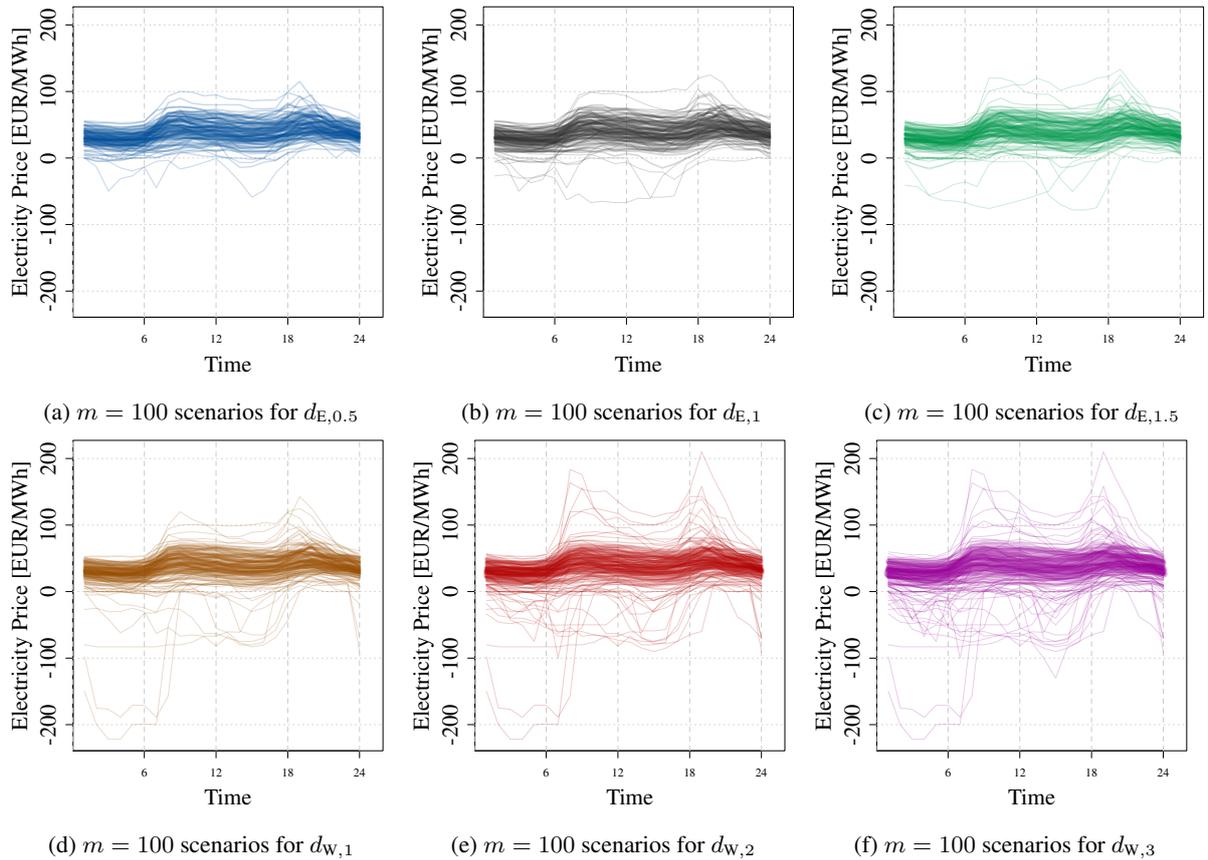

        \begin{subfigure}[b]{0.325\textwidth}
\centering
\resizebox{.999\textwidth}{!}{\input{fig/price_sel_cdn=100_1.tex} }
        \caption{$m=100$ scenarios for $d_{\text{E},0.5}$}
        \label{fig_paths_1}
    \end{subfigure}
        \begin{subfigure}[b]{0.325\textwidth}
\centering
\resizebox{.999\textwidth}{!}{\input{fig/price_sel_cdn=100_2.tex} }
        \caption{$m=100$ scenarios for $d_{\text{E},1}$}
        \label{fig_paths_2}
    \end{subfigure}
        \begin{subfigure}[b]{0.325\textwidth}
\resizebox{.999\textwidth}{!}{\input{fig/price_sel_cdn=100_3.tex} }
        \caption{$m=100$ scenarios for $d_{\text{E},1.5}$}
        \label{fig_paths_3}
    \end{subfigure}
        \begin{subfigure}[b]{0.325\textwidth}
\centering
\resizebox{.999\textwidth}{!}{\input{fig/price_sel_cdn=100_4.tex} }
        \caption{$m=100$ scenarios for $d_{\text{W},1}$}
        \label{fig_paths_4}
    \end{subfigure}
        \begin{subfigure}[b]{0.325\textwidth}
\centering
\resizebox{.999\textwidth}{!}{\input{fig/price_sel_cdn=100_5.tex} }
        \caption{$m=100$ scenarios for $d_{\text{W},2}$}
        \label{fig_paths_5}
    \end{subfigure}
        \begin{subfigure}[b]{0.325\textwidth}
\centering
\resizebox{.999\textwidth}{!}{\input{fig/price_sel_cdn=100_6.tex} }
        \caption{$m=100$ scenarios for $d_{\text{W},3}$}
        \label{fig_paths_6}
    \end{subfigure}
\caption{Illustration of randomly selected scenarios $m=100$ scenarios out of $n=3653$ by the energy distances $d_{\text{E},0.5}$, $d_{\text{E},1}$ and $d_{\text{E},2}$, and Wasserstein distances $d_{\text{W},1}$, $d_{\text{W},2}$ and $d_{\text{W},3}$. }
\label{fig_paths}
\end{figure}

Additionally, we may deduce that the Wasserstein distance tends to replicate the tail and outlier properties better than energy distance. The latter is preferable for center properties which involve the full distribution, like mean and variance characteristics. This gets even clearer if we look at Figure \ref{fig_paths}. It shows the 100 selected paths out of the 3653 paths for all six metrics. We observe that reduced scenario sets of the Wasserstein distances $d_{W,p}$ include more outlier trajectories than the energy distance reduction, and even more with increasing $p$. Consequently, the support of the full distribution is much better recovered. For $d_{W,2}$ and $d_{W,3}$, even the overall minimum and maximum (see \ref{tab_summary}) is identified correctly. This comes at some cost of worse approximation in the non-tail and non-outlier properties.
Thus, if scenario reduction is applied to stochastic optimization, the choice of the distance should likely depend on the stochastic properties that are crucial in the optimization. If mainly the tail and outlier events are relevant, then the Wasserstein distance is likely the first choice, otherwise the energy distance seems to be preferable.


\section{Discussion and Conclusion}
We illustrate that the energy distance is a suitable distance for ensemble and scenario reduction problems. It tends to provide better statistical/stochastic approximation properties than the popular Wasserstein distance which is vastly used in scenario reduction applications. Next to important characteristics like the approximation in the first two moments, this holds particularly for path-dependency properties. They are relevant in many applications,
especially for those in stochastic programs in energy systems like storage, maintenance, and power trading optimization. 

However, one drawback is that the reduction methods based on the energy distance is computationally slightly more demanding than the Wasserstein distance. This holds especially for the scenario reduction. Here, the Wasserstein distance has a simple explicit solution based on the redistribution rule which is linked to transportation problems. The energy distance on the other hand requires to solve a quadratic program with linear constraints. 
Due to the structure of the optimization problem \cite{vavasis1992local} proves the existence of 
an algorithm that solves the quadratic program in quadratic time with respect to reduced scenario size. 
To our knowledge, such an algorithm is still unknown, further research should investigate potential optimization algorithms.
Nevertheless, both the Wasserstein and energy distance lead to reduction problems that are NP-hard and require advanced search algorithms even for moderately sized problems. Sophisticated selection methods that work for the Wasserstein metrics can be applied to the energy distance as well, see e.g. \cite{li2019fast}.

From the methodological point, the research can go in different directions. 
We only discussed briefly the impact of the parameter $p$ in the energy distance $d_{\text{E},p}$ and Wasserstein distance $d_{\text{W},p}$, and focused on the standard choices for $p=1$. However, further investigations on the impact of $p$ might be suitable. For heavy tailed data, the energy distance with $p<1$ should provide more stable results \cite{szekely2013energy, szekely2017energy}.

Moreover, the energy distance can be generalized to a maximum mean discrepancy (MMD) concept using kernels (see e.g. \cite{borgwardt2006integrating, szekely2017energy}). Then the distance measure is given by
$ \E \kappa(\bsX-\bsY)- \frac{1}{2} \E \kappa( \bsX-\bsX' )  -\frac{1}{2} \E \kappa(\bsY-\bsY') $ for suitable kernels $\kappa$. However, 
as noted by \cite{sejdinovic2013equivalence} there is no superior kernel $\kappa$ that serves well in all situations. 
Furthermore, popular kernels from machine learning like Gaussian 
or Laplacian kernels suffer disadvantages in high dimensions and do not satisfy scale equivariance, see \cite{szekely2017energy}.
Thus, the standard energy distance (see \eqref{eq_energy_dist}) seems to be a adequate natural candidate for scenario and ensemble reduction.
Furthermore, the Sinkhorn divergence provides an interesting combination between the energy and Wasserstein distance, see e.g. \cite{genevay2017learning}. 
Here, further investigations with respect to scenario reduction problems might be useful to combine the beneficial properties from both worlds. However, advantage of the low computational complexity of the Wasserstein distance for scenario reduction will be lost.


\bibliographystyle{apalike}
\bibliography{scenario_red}

\end{document}